\documentclass[preprint,12pt]{elsarticle}

\usepackage[T1]{fontenc}
\usepackage[breaklinks,colorlinks]{hyperref}
\usepackage{graphicx}
\usepackage{multirow}
\usepackage{tikz}
\usepackage{tikz-3dplot}
\usepackage{pgfplots}
\pgfplotsset{width=8cm,compat=1.14}
\usepackage{amsmath}
\usepackage{booktabs}

\usetikzlibrary{decorations,decorations.markings,decorations.text}
\usetikzlibrary{calc}
\usetikzlibrary{shadings}
\usepackage{subcaption} \usepackage[normalem]{ulem}
\usepackage{xcolor}
\usepackage{soul}
\useunder{\uline}{\ul}{}

\usepackage[capitalize]{cleveref}
\crefname{section}{Sec.}{Secs.}
\Crefname{section}{Section}{Sections}
\Crefname{table}{Table}{Tables}
\crefname{table}{Tab.}{Tabs.}
\usepackage{amssymb}\usepackage{pifont}\newcommand{\cmark}{\ding{51}}\newcommand{\xmark}{\ding{55}}\newcommand{\R}{\mathbb{R}}

\definecolor{color1}{rgb}{0.466, 0.929, 0.466} \definecolor{color2}{rgb}{0.219, 0.658, 0.219} \definecolor{color3}{rgb}{0.129, 0.388, 0.129} \definecolor{paleorange}{RGB}{255,204,153} 
\definecolor{cvcred}{HTML}{FF0000}  
\definecolor{cvcorange}{HTML}{FFD200} \definecolor{cvcblue}{HTML}{3d66fc} \definecolor{cvcgreen}{RGB}{52, 137, 43} \usepackage{amssymb}
\usepackage{amsmath}

\journal{Pattern Recognition}

\begin{document}

\begin{frontmatter}

\title{The OCR Quest for Generalization:\\Learning to recognize low-resource alphabets with model editing}

\author{Adrià Molina,
        Oriol Ramos Terrades,
        Josep Lladós}

\affiliation{
  organization={Centre de Visió per Computador (CVC)},
}

\affiliation{
  organization={Computer Science Department, Universitat Autònoma de Barcelona},
}

\begin{abstract}
Achieving robustness in recognition systems across diverse domains is crucial for their practical utility. While ample data availability is usually assumed, low-resource languages, such as ancient manuscripts and non-western languages, tend to be kept out of the equations of massive pretraining and foundational techniques due to an under representation. In this work, we aim for building models which can generalize to new distributions of data, such as alphabets, faster than centralized fine-tune strategies. For doing so,  we take advantage of the recent advancements in model editing to enhance the incorporation of unseen scripts (low-resource learning). In contrast to state-of-the-art meta-learning, we showcase the effectiveness of domain merging in sparse distributions of data, with agnosticity of its relation to the overall distribution or any other prototyping necessity. Even when using the same exact training data, our experiments showcase significant performance boosts in \textbf{transfer learning} to new alphabets and \textbf{out-of-domain evaluation} in challenging domain shifts, including historical ciphered texts and non-Latin scripts. This research contributes a novel approach into building models that can easily adopt under-represented alphabets and, therefore, enable document recognition to a wider set of contexts and cultures.
\end{abstract}

% \begin{highlights}
% \item We demonstrate that task arithmetic serves as an effective mechanism to facilitate the integration of low-resource and non-Latin languages into modern OCR-based reading systems, improving inclusivity in document understanding tasks and reading systems.
% \item We show that the proposed approach also exhibits some generalization capabilities, as evidenced by its performance in numerous out-of-domain evaluations across diverse and challenging data distributions.
% \item We substantiate our claims through an extensive and reproducible experimental protocol involving over 100 trained models, evaluated on more than 20 datasets representing a broad spectrum of scripts and domains. All trained models are publicly available to foster transparency and community adoption.
% \end{highlights}


\end{frontmatter}

\section{Introduction}
\label{sec:intro}
Document Intelligence (DI) encompasses advanced AI capabilities designed to interpret various types of documents. At the core of DI applications are reading systems that process text in both modalities: typewritten text, using Optical Character Recognition (OCR), and handwritten text, through Intelligent Character Recognition (ICR). This distinction is made while maintaining language agnosticity. However, in the Deep Learning era, the reliance on data distributions tends to bias models, leading to improved performance in dominant languages.  While some recent studies suggest that the incorporation of low-resource and non-Western alphabets can be addressed through large-scale pre-training regimes \cite{shu2024transcending,zhang2024mc2}, other researchers highlight the continued underrepresentation of such languages in modern large language models (LLMs) \cite{lin2024one,smith2024standard,tran2025beyond}. This discrepancy is primarily attributed to the optimization strategies employed in model training, which tend to favor the standardization of English as the dominant language in most large-scale datasets.

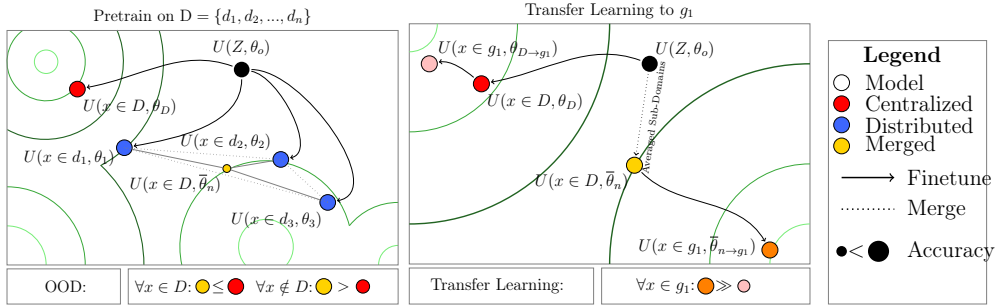
\begin{figure}[t]
    \centering
    \begin{subfigure}{0.39\textwidth}
        \centering
        \resizebox{\textwidth}{!}{

\begin{tikzpicture}

    \coordinate (B) at (5,5);
    \coordinate (R) at (0.8,4.5);
    \coordinate (BL1) at (2,3);
    \coordinate (BL2) at (6,2.7);
    \coordinate (BL3) at (7.2, 1.6);
    \coordinate (O) at ($(BL1)!0.333!(BL2)!0.333!(BL3)$);

    \def\center{(0,5.2)}
    \draw[color1, line width=0.15pt] \center ++(0:0.3) arc (0:360:0.3);
    \draw[color2, line width=0.5pt] \center ++(-233:1) arc (-233:53:1);
    \draw[color3, line width=0.8pt] \center ++(-119.8:2) arc (-119:23.5:2);
    \draw[color3, line width=0.8pt] \center ++(-62.3:3) arc (-62.3:15.5:3);

    \draw[color2, line width=0.15pt] ($(O) +(1, -2)$) ++(192.3:2.2) arc (192.3:14:2.2);
    \draw[color1, line width=0.15pt] ($(O) +(1, -2)$) ++(222:0.7) arc (222:-41:0.7);
    \def\centerright{(9,0)}
    \draw[color2, very thin]  \centerright ++(141:1.6) arc (141:90:1.6);
    \draw[color1, very thin]  \centerright ++(90:0.5) arc (90:180:0.5);

    \def\centerleft{(-1,0)}
    \draw[color1, line width=0.005pt] \centerleft ++(0:1) arc (0:90:1);
    \draw[color2, line width=0.15pt]  \centerleft ++(0:2) arc (0:90:2);
    \draw[color3, line width=0.8pt]  \centerleft ++(0:3.5) arc (0:46.8:3.5);

    \draw[gray, very thin] (-1,0) rectangle (9,6);
    \draw[gray, very thin] (2.1,-0.1) rectangle (9,-1);
    \draw[gray, very thin] (-1,-0.1) rectangle (1.9,-1);
    
    \node at (0.5, -0.55) {OOD:}; \coordinate (CAPTION) at (3, -0.55);
    
    \node at (CAPTION) {$\forall x \in D$:};
    \fill[cvcorange, draw=black] ($(CAPTION) + (1, 0)$) circle (.17);
    \node at ($(CAPTION) + (1, 0) + (0.4, 0)$) {$\leq$};
    \fill[cvcred, draw=black] ($(CAPTION) + (1, 0) + (0.4, 0) + (0.45, 0)$) circle (.2);

    \coordinate (CAPTION2) at ($(CAPTION) + (1, 0) + (0.4, 0) + (0.45, 0) + (0.45, 0) + (0.8, 0)$);

    \node at (CAPTION2) {$\forall x \notin D$:};
    \fill[cvcorange, draw=black] ($(CAPTION2) + (1, 0)$) circle (.2);
    \node at ($(CAPTION2) + (1, 0) + (0.5, 0)$) {$>$};
    \fill[cvcred, draw=black] ($(CAPTION2) + (1, 0) + (0.4, 0) + (0.6, 0)$) circle (.17);

    \node[above] at (4,6) {Pretrain on $\text{D}=\{d_1, d_2, ..., d_n\}$};

    \fill[black] (B) circle (.2);
    \fill[cvcred, draw=black] (R) circle (.2);
    \fill[cvcblue, draw=black] (BL1) circle (.2);
    \fill[cvcblue, draw=black] (BL2) circle (.2);
    \fill[cvcblue, draw=black] (BL3) circle (.2);
    \fill[cvcorange, draw=black] (O) circle (.1);
    
    \node[above] at (5,5.2) {$U(Z, \theta_o)$};
    \node[below right] at ($(R) + (0, -0.1)$) {$U(x\in D, \theta_D)$};

    \node[below left] at ($(BL1) + (-0.1, 0.1)$) {$U(x \in d_1, \theta_1)$};
    \node[above left] at ($(BL2) + (-0.1, 0.1)$) {$U(x \in d_2, \theta_2)$};
    \node[below left] at ($(BL3) + (0, -0.1)$) {$U(x \in d_3, \theta_3)$};
    \node[below left] at (O) {$U(x \in D, \overline{\theta}_n)$};

    \draw [black, ->, shorten <= 0.25cm, shorten >= 0.25cm] (B) to[out=160,in=10] (R);
    \draw [black, ->, shorten <= 0.25cm, shorten >= 0.25cm] (B) to[out=270,in=10] (BL1);
    \draw [black, ->, shorten <= 0.25cm, shorten >= 0.25cm] (B) to[out=-10,in=10] (BL2);
    \draw [black, ->, shorten <= 0.25cm, shorten >= 0.25cm] (B) to[out=0,in=10] (BL3);

    \draw [gray, dotted] (BL1) -- (BL2);
    \draw [gray, dotted] (BL2) -- (BL3);
    \draw [gray, dotted] (BL3) -- (BL1);
    \draw [gray, thin, shorten <= 0.1cm, shorten >= 0.2cm] (O) -- (BL1);
    \draw [gray, thin, shorten <= 0.1cm, shorten >= 0.2cm] (O) -- (BL2);
    \draw [gray, thin, shorten <= 0.1cm, shorten >= 0.2cm] (O) -- (BL3);

    \end{tikzpicture}}
        \end{subfigure}
        \begin{subfigure}{0.39\textwidth}
        \centering
        \resizebox{\textwidth}{!}{\begin{tikzpicture}

    \draw[gray, very thin] (-1,0) rectangle (9,6);
    \draw[gray, very thin] (3.8,-0.1) rectangle (9,-1);
    \draw[gray, very thin] (-1,-0.1) rectangle (3.6,-1);

    \def\center{(-1,6)}
    \draw[color1, line width=0.01pt] \center ++(0:0.7) arc (0:-90:0.7);
    \draw[color2, line width=0.1pt] \center ++(0:2.7) arc (0:-90:2.7);
    \draw[color3, line width=1pt] \center ++(0:5) arc (0:-90:5);

    \def\centerleft{(9,0)}
    \draw[color1, line width=0.01pt] \centerleft ++(180:1) arc (180:90:1);
    \draw[color2, line width=0.1pt] \centerleft ++(180:2.5) arc (180:90:2.5);
    \draw[color3, line width=1pt] \centerleft ++(180:5) arc (180:90:5); 

    \node at (1.2, -0.55) {Transfer Learning:}; \coordinate (CAPTION) at (5.4, -0.55);
    
    \node at (CAPTION) {$\forall x \in g_1$:};
    \fill[orange, draw=black] ($(CAPTION) + (1, 0)$) circle (.2);
    \node at ($(CAPTION) + (1, 0) + (0.45, 0)$) {$\gg$};
    \fill[pink, draw=black] ($(CAPTION) + (1, 0) + (0.4, 0) + (0.55, 0)$) circle (.15);

    \node[above] at (4,6) {Transfer Learning to $g_1$};    
    \coordinate (B) at (5,5);
    \coordinate (R) at (0.8,4.5);
    \coordinate (BL1) at (2,3);
    \coordinate (BL2) at (6,2.7);
    \coordinate (BL3) at (7.2, 1.6);
    \coordinate (O) at ($(BL1)!0.333!(BL2)!0.333!(BL3)$);

    \coordinate (P) at (-0.5,5);
    \coordinate (Y) at (8, 0.35);

    \fill[black] (B) circle (.2);
    \fill[cvcred, draw=black] (R) circle (.2);
    \fill[cvcorange, draw=black] (O) circle (.2);
    \fill[pink, draw=black] (P) circle (.2);
    \fill[orange, draw=black] (Y) circle (.2);

    \node[above right] at (B) {$U(Z, \theta_o)$};
    \node[below right] at ($(R) + (0, -0.1)$) {$U(x\in D, \theta_D)$};
    \node[above right] at ($(P) + (0.08, 0)$) {$U(x\in g_1, \theta_{D \rightarrow g_1} )$};
    \node[left] at ($(Y) + (-0.2, 0.1)$) {$U(x\in g_1, \overline{\theta}_{n \rightarrow g_1} )$};

    \node[below left] at (O) {$U(x \in D, \overline{\theta}_n)$};

    \pgfmathanglebetweenpoints{\pgfpointanchor{O}{center}}{\pgfpointanchor{B}{center}}
    \let\angle\pgfmathresult
    \draw [black, dotted, <-, shorten <= 0.25cm, shorten >= 0.25cm] (O) -- (B) node[rotate=\angle] at ($(O) + (0.5, 1.2)$) {\fontsize{7}{8}\selectfont{Averaged Sub-Domains}};

    \draw [black, ->, shorten <= 0.25cm, shorten >= 0.25cm] (B) to[out=160,in=10] (R);
    \draw [black, ->, shorten <= 0.25cm, shorten >= 0.25cm] (R) to[out=140,in=10] (P);
    \draw [black, ->, shorten <= 0.25cm, shorten >= 0.25cm] (O) to[out=-45,in=120] (Y);

    \end{tikzpicture}}
        \end{subfigure}
    \begin{subfigure}{0.2\textwidth}
    \centering
    \resizebox{\textwidth}{!}{\def\startingx{0.25} 
\begin{tikzpicture}
    \draw[gray, very thin] (0,0) rectangle (3,5);

    \node at (1.4,4.7) {\textbf{Legend}};

    \draw[black] (\startingx,4.2) circle (0.15);
    \node[anchor=west, text width=2cm, align=left] at ({\startingx + 0.3},4.2) {Model};

    \fill[cvcred, draw=black] (\startingx,3.8) circle (0.15);
    \node[anchor=west, text width=2cm, align=left] at ({\startingx + 0.3},3.8) {Centralized};

    \fill[cvcblue, draw=black] (\startingx,3.4) circle (0.15);
    \node[anchor=west, text width=2cm, align=left] at ({\startingx + 0.3},3.4) {Distributed};

    \fill[cvcorange, draw=black] (\startingx,3.0) circle (0.15);
    \node[anchor=west, text width=2cm, align=left] at ({\startingx + 0.3},3.0) {Merged};

    \draw[->, thick] (\startingx,2.4) -- ({\startingx + 1.0},2.4);
    \node[anchor=west, text width=2cm, align=left] at ({\startingx + 1.1},2.4) {Finetune};

    \draw[dotted, thick] (\startingx,1.8) -- ({\startingx + 1.0},1.8);
    \node[anchor=west, text width=2cm, align=left] at ({\startingx + 1.1},1.8) {Merge};

    \fill[black] (\startingx,1.0) circle (0.10);
    \node at ({\startingx + 0.28},1.0) {$<$};
    \fill[black] ({\startingx + 0.7},1.0) circle (0.20);
    \node[anchor=west, text width=2cm, align=left] at ({\startingx + 1.1},1.0) {Accuracy};
\end{tikzpicture}
}
\end{subfigure}

    \caption{Summary of the two main contributions of this work. Given a \textbf{shared model initialization}, generalization (left) is improved with respect to \colorbox{red!30}{centralized pre-training} (red) by using model merging from training in a \colorbox{blue!30}{distributed manner} (blue). \colorbox{yellow!30}{Our approach} (yellow) outperforms centralized regimes in out-of-distribution data ($x \notin D$). This is also applied to transfer learning (right), where \colorbox{orange!30}{our method} (orange) proves advantageous in incorporating low-resource data from historical ciphers and non-western languages ($x \in g_i$). Notation is to be found in later sections.}

\label{fig:training_regimes}
\end{figure}

Acquiring large annotated corpora for certain languages is significantly more challenging than for plain Latin alphabets and, in some cases, outright prohibitive. This challenge becomes even more pronounced in the context of \textbf{historical document analysis}, where scarce and unique data make annotation especially costly.

Due to the above reasons, practitioners have leveraged few-shot learning \cite{huang2021zero,souibgui2022few} and prototyping techniques \cite{roy2011word,vlachou2024interpretable,yan2021primitive} to develop feature extractors capable of generalizing effectively in subsequent low-resource contexts. However, such approaches may cause unnecessary computational expenses and loss of generality due to operating directly on the data at hand.

In the increasingly complex paradigm of modern machine learning, computational burden is prohibitive in many contexts. The alternative for prototyping usually lies in expensive centralized regimes (i.e. large-scale pretraining), which have been flagged due to the privacy issues of accumulating data in a single deep learning model \cite{carlini2023extracting,zhang2020fedocr}. This is particularly important in historical document analysis, where centralization raises, additionally, preservation concerns \cite{jaillant2022unlocking,lawrence2016privacy,solterodecentralized}.

Some historical archives have highlighted the importance of preserving documents close to their place of origin
\cite{planes2013arxius,brothman2001past,ketelaar2005sharing}, as it guarantees that their original context, authenticity, and cultural significance are maintained \cite{olliff2021distributed}. This, in turn, aligns with the idea of achieving \textbf{generalization within a distributed setup}, where one can generalize while using a set of distributed "nodes" (i.e. independent processes, not necessarily on the same computer or within the same organization) and, thus, without the need of storing historical documents away from their original source.

As shown in \figureautorefname~\ref{fig:training_regimes}, this work leverages model merging algorithms to develop reading systems that can efficiently learn new alphabets—such as those in historical ciphered texts or low-resource languages—by learning robust features. Our approach enables data-efficient incorporation of non-Latin scripts without requiring centralized data (i.e., domains can be trained independently).

Unlike \cite{nichol2018first}, we adopt a meta-learning-inspired strategy to handle target datasets $g$ that differ significantly from the original distribution. We show that this improves adaptability in scenarios involving disjoint data distributions $d_n$.

This paper contributes to the document analysis and pattern recognition community through the following three key aspects:

\begin{enumerate}

\item It is empirically demonstrated that \textbf{meta-learned reading systems exhibit robust feature representations} through \textit{out-of-domain} evaluations. In contrast to previous approaches, we propose a practical scenario by using a disjoint set of data distributions.

\item Fine-tuning is conducted on these generalized models to demonstrate the effectiveness of \textbf{distributed training regimes in \textit{domain adaptation} in low-resource alphabets}.

\item A unified evaluation baseline is established across 20 major OCR datasets - including handwritten, scene, printed, historical, ciphered, and cross-lingual texts. \textbf{Over 100 trained and fine-tuned models} (baselines, per-language/cipher ablations, out-of-domain merges) are provided to the Document Analysis community. Their performance is documented in a publicly available model card\footnote{\url{https://eaudedata.github.io/ODAOCR/model_card/model_card.html}}, where all model weights can be accessed.

\end{enumerate}

In this manner, we demonstrate the effectiveness of a method that offers a \textbf{decentralized approach to historical and low-resource document analysis}, delivering superior performance compared to centralized large-scale pre-training without necessarily relying on prototyping or any kind of data augmentation technique. 

This work is structured as follows: Section \ref{sec:sota} reviews Low-Resource OCR literature. Section \ref{sec:method} introduces our method, while Section \ref{sec:exp} details the experimental setup. Results are presented in Section \ref{sec:res}. Finally, Section \ref{sec:conc} summarizes key findings and insights for OCR practitioners.

\section{Related Work}
\label{sec:sota}
Methods for data-efficient Optical Character Recognition (OCR) have long been explored within the document analysis community. A pioneering effort by Taghv \textit{et al.} \cite{taghva1994expert} introduced an ensemble of heuristics to enhance OCR predictions (post-OCR processing) when encountering potential prediction errors. In the deep learning era, Wick \textit{et al.} \cite{wick2021one} addressed the alignment problem in Connectionist Temporal Classification (CTC) models, introducing a method to facilitate ensembling of confidence scores but requiring simultaneous training across multiple domains. More recently, Seuret \textit{et al.} \cite{seuret2023combining} proposed a weighted averaging approach proportional to domain classification confidence, offering a flexible framework but still relying on predefined domain enumeration.

As noted in \cite{souibgui2022few}, we can find examples of few-shot tasks as a semi-supervised pseudo-labeling task. These approaches range from pseudo-labeling for data augmentation \cite{szummer2001partially,weston2008deep} to some curriculum approaches \cite{choi2019pseudo,lee2013pseudo}. Authors at \cite{souibgui2022few}, utilize a similar prototyping technique to alleviate the scarcity of handwritten document data. In this work, we aim to propose an optimization technique to improve recognition in low-resource scenarios, as the ones to be found in \cite{souibgui2022few}, without any additional mechanism than gradient descent algorithms. More specifically, we propose adapting narrow domains (historical ciphers and other low-resource alphabets) by generalizing on a wider set of tasks. 

Given the growing emphasis in pattern recognition literature on developing systems that can seamlessly generalize to diverse environments, as introduced by Ilharco \textit{et al.} \cite{ilharco2023editing}, this work builds upon the emerging field of task arithmetic to construct robust OCR systems. Unlike the post-hoc generalization approach explored in \cite{ilharco2023editing}, some studies integrate task arithmetic as a core component of the optimization process \cite{nichol2018first}. This approach, commonly referred to as meta-learning, has been particularly effective in few-shot learning scenarios for acquiring new concepts within a shared domain. However, to the best of our knowledge, no prior work has addressed the challenges of applying meta-learning to contexts involving strong distribution shifts, such as alphabet shifts across different domains. In the following sections, we will describe the characteristics of these partitions.

\begin{figure}[t]
    \centering
    \begin{subfigure}{0.32\textwidth}
        \centering
        \resizebox{\textwidth}{!}{
        \begin{tikzpicture}
\draw [thick]  plot[smooth, tension=.7] coordinates {(-4,2.5) (-3.5,2.7) (-3,3) (-2.5,3) (-2,2.8) (-1.5,2.6) (-1,2.3) (-0.8,2.5) (-0.5,1.5) (0,1) (0.5,0) (0,-1) (-0.5,-2) (-1,-2.2) (-1.5,-2.5) (-2,-2.7) (-2.5,-2.8) (-3,-2.8) (-3.5,-2.6) (-4,-2) (-4.5,-1) (-5,0) (-4.5,1) (-4,2.5)};
\draw [fill=black!20!green!80!white] plot[smooth cycle, tension=.7] coordinates {
     (-4,1.5)  (-3,-1) (-2.5,-1.7) (-1.5,-2) (-0.5,-1) (-0.5,0) (-1,1.5) (-2.5,2.5) 
};

\draw [fill=black!20!color1!60!white] plot[smooth cycle, tension=.7] coordinates {
    (-1.5,0) (-2,0.5) (-2.5,1) (-1.5,1.5) (-0.5,1) (-0.5,0.5) (-0.8,0) (-1.5,-0.5)
};

\draw [fill=black!20!color3!60!white] plot[smooth cycle, tension=.7] coordinates {
    (-0.5, -1.8) (-1.5, -0.6) (-2.3, -1.6)  
};

\node at (-3.8,-0.7) {$x \notin D$};
\node at (-3,1.7) {$x \in D$};

\node at (-1.5, -1.5) {$x \in d_1$};
\node at (-1.2,1) {$x \in d_2$};

\coordinate (R) at (-2.3, 0.1);
\coordinate (BL1) at (-1.3, 0.5);
\coordinate (BL2) at (-1.1, -1.1);

\fill[red, draw=black] (R) circle (.1);
\node[below] at ($(R) + (-0.1, 0)$) {$\theta_D$};

\fill[cyan, draw=black] (BL1) circle (.1);
\node[below right] at ($(BL1) + (0, 0)$) {$\theta_{d_2}$};

\fill[cyan, draw=black] (BL2) circle (.1);
\node[left] at ($(BL2) + (0, 0)$) {$\theta_{d_1}$};

\end{tikzpicture}
        }
        \caption{Training $x \in D$}
        \label{fig:domains_a}
    \end{subfigure}
        \begin{subfigure}{0.32\textwidth}
        \centering
        \resizebox{\textwidth}{!}{

\begin{tikzpicture}
\draw [thick]  plot[smooth, tension=.7] coordinates {(-4,2.5) (-3.5,2.7) (-3,3) (-2.5,3) (-2,2.8) (-1.5,2.6) (-1,2.3) (-0.8,2.5) (-0.5,1.5) (0,1) (0.5,0) (0,-1) (-0.5,-2) (-1,-2.2) (-1.5,-2.5) (-2,-2.7) (-2.5,-2.8) (-3,-2.8) (-3.5,-2.6) (-4,-2) (-4.5,-1) (-5,0) (-4.5,1) (-4,2.5)};
\draw [fill=black!20!green!80!white] plot[smooth cycle, tension=.7] coordinates {
     (-4,1.5)  (-3,-1) (-2.5,-1.7) (-1.5,-2) (-0.5,-1) (-0.5,0) (-1,1.5) (-2.5,2.5) 
};

\draw [fill=black!20!color1!60!white] plot[smooth cycle, tension=.7] coordinates {
    (-1.5,0) (-2,0.5) (-2.5,1) (-1.5,1.5) (-0.5,1) (-0.5,0.5) (-0.8,0) (-1.5,-0.5)
};

\draw [fill=black!20!color3!60!white] plot[smooth cycle, tension=.7] coordinates {
    (-0.5, -1.8) (-1.5, -0.6) (-2.3, -1.6)  
};

\node at (-3.8,-0.7) {$x \notin D$};
\node at (-3,1.7) {$x \in D$};

\coordinate (R) at (-2.3, 0.1);
\coordinate (BL1) at (-1.3, 0.5);
\coordinate (BL2) at (-1.1, -1.1);

\fill[red, draw=black] (R) circle (.1);
\node[below] at ($(R) + (-0.1, 0)$) {$\theta_D$};

\fill[cyan, draw=black] (BL1) circle (.1);

\fill[cyan, draw=black] (BL2) circle (.1);

\coordinate (mid) at ($(BL1)!0.5!(BL2)$);
\draw [black, dashed] (BL1) -- (BL2);
\fill [orange, draw=black] (mid) circle (.1);

\node [below right] at (mid) {$\bar{\theta}_n$};

\end{tikzpicture}}
                \caption{Equation \ref{eq:metode}}

        \label{fig:domains_b}
    \end{subfigure}
    \begin{subfigure}{0.32\textwidth}
        \centering
        \resizebox{\textwidth}{!}{\begin{tikzpicture}
\draw [thick]  plot[smooth, tension=.7] coordinates {(-4,2.5) (-3.5,2.7) (-3,3) (-2.5,3) (-2,2.8) (-1.5,2.6) (-1,2.3) (-0.8,2.5) (-0.5,1.5) (0,1) (0.5,0) (0,-1) (-0.5,-2) (-1,-2.2) (-1.5,-2.5) (-2,-2.7) (-2.5,-2.8) (-3,-2.8) (-3.5,-2.6) (-4,-2) (-4.5,-1) (-5,0) (-4.5,1) (-4,2.5)};
\draw [fill=black!20!green!80!white]  plot[smooth cycle, tension=.7] coordinates {
     (-4,1.5)  (-3,-1) (-2.5,-1.7) (-1.5,-2) (-0.5,-1) (-0.5,0) (-1,1.5) (-2.5,2.5) 
} ;

\draw [fill=black!20!color1!60!white] plot[smooth cycle, tension=.7] coordinates {
    (-1.5,0) (-2,0.5) (-2.5,1) (-1.5,1.5) (-0.5,1) (-0.5,0.5) (-0.8,0) (-1.5,-0.5)
};

\draw [fill=black!20!color3!60!white] plot[smooth cycle, tension=.7] coordinates {
    (-0.5, -1.8) (-1.5, -0.6) (-2.3, -1.6)  
};

\node at (-3.8,-0.7) {$x \notin D$};
\node at (-3,1.7) {$x \in D$};

\coordinate (R) at (-2.3, 0.1);
\coordinate (BL1) at (-1.3, 0.5);
\coordinate (BL2) at (-1.1, -1.1);

\fill[red, draw=black] (R) circle (.1);
\fill[cyan, draw=black] (BL1) circle (.1);
\fill[cyan, draw=black] (BL2) circle (.1);

\coordinate (mid) at ($(BL1)!0.5!(BL2)$);
\draw [black, dashed] (BL1) -- (BL2);
\fill [orange, draw=black] (mid) circle (.1);

\draw [black, |<->|] ($(mid) + (0.15, 0)$) -- ($(mid) + (0.73, 0)$);

\draw [black, |<->|] ($(R) + (-0.15, 0)$) -- ($(R) + (-1.24, 0)$);

\end{tikzpicture}}
                \caption{ Evaluate $ x \notin D$}

        \label{fig:domains_c}
    \end{subfigure}
    \caption{Model sampling method for out-of-domain generalization, where models $\theta$ are: (a) optimized towards a domain $D$ or $d_n$, (b) task arithmetic averaging, which consists in creating a new model as an average of the previous ones, and (c)  is then evaluated out-of-domain. }
    \label{fig:domains}
\end{figure}
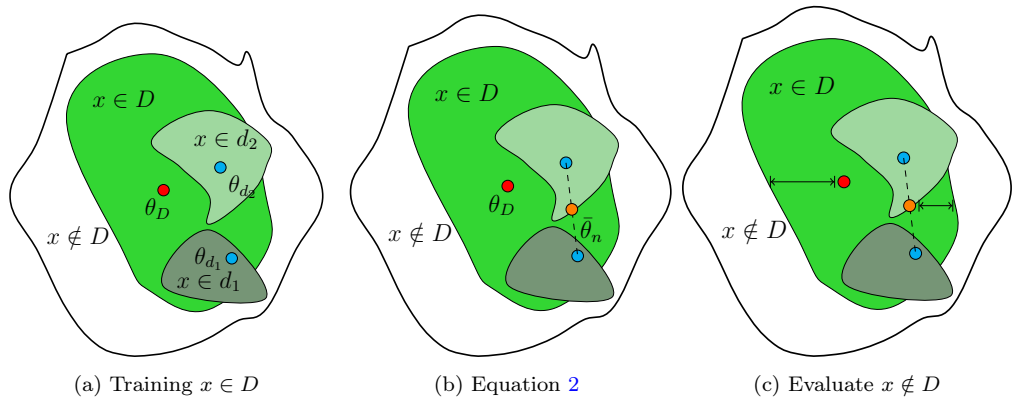

\section{Generalization Method}
\label{sec:method}
In this section, we present our strategy for sampling robust models. First, we utilize the averaging of expert Latin models for out-of-distribution evaluation (see Figure~\ref{fig:training_regimes}, orange). Second, building upon these sampled models, we provide practitioners with a straightforward approach to creating models that significantly enhance performance in new low-resource domains (see Figure~\ref{fig:training_regimes}, yellow), such as new languages or ciphered scripts. 
\subsection{Notation}
\label{sec:notation}
Before presenting a full definition of the employed method, we first summarize the basic notation. For further details on the mathematical formalism, readers may refer to \cite{nichol2018first} and \cite{ortizjimenez2023task}, as this work builds upon similar concepts and ideas.

The basic elements constituting our method are: 

\begin{itemize}  
    \item A fully centralized dataset $D$, consisting of $N$ different domains $d_n$.  
    \item A set of out-of-distribution data points $x \notin D$ and a new domain $g$ for transfer learning.  
    \item The model weights, represented as a single vector $\theta$.  
    \item An update function $U^k$, which takes a model and a dataset as input and returns a new model after performing $k$ optimization steps on the given dataset.  
    \item A number of epochs $T$, during which the function $U^k(d, \theta)$ is iteratively applied $T$ times.  
\end{itemize}
\subsection{Preliminaries}
In this section, we aim to establish the necessary preliminaries to unify the notation used throughout this work, given the diverse research directions that converge into the \textit{FedAvg} algorithm~\cite{mcmahan2023communication}, it is necessary to establish equivalences between Reptile~\cite{nichol2018first}, task arithmetic~\cite{ilharco2023editing} and \textit{FedAvg}, as all of the stated methods induce an interpretation of a common algorithm from different perspectives.

Intuitively, this methods train different models on different data distributions (see blue dots in Figure~\ref{fig:training_regimes}). While the data separation usually consists on disjoint distributions, in the case of Reptile~\cite{nichol2018first} the data distributions are randomly sampled from an original dataset where one category is left out for evaluation; which stabilizes the training process by merging models that did not diverge too much. This models are then arithmetically merged~\cite{ilharco2023editing} for conducting posterior training stages. In our case, we go one step further and utilize this FedAvg models to conduct fine-tunning in a domain adaptation set-up (see yellow dot in Figure~\ref{fig:training_regimes}).

First, the \textit{FedAvg} algorithm~\cite{mcmahan2023communication} initializes a network with a set of parameters $\theta$ and data $D$ that is disjointly distributed among $N$ clients $\{d_1, d_2, \ldots, d_n\}$. In real-world applications, each client represents a different dataset of a common task. Once the different partitions are defined, an update function $U(D, \theta)$ is introduced, which consists of $k$ stochastic optimization steps (such as Adam~\cite{kingma2017adam}) using the data partition $d_n$. This update process results in $U^k( d_n; \theta)$. A learning optimization process, defined in Equation \ref{eq:fedAvg}, is then repeated $T$ times. This training regime is ruled by the number of stochastic optimization steps $k$ and the number of aggregation steps $T$.

\begin{equation}
    \theta_{t+1} \leftarrow \theta_t + \frac{1}{N}\sum^{N}_{n} \left[ U^k( d_n; \theta_t) - \theta_t \right]
    \label{eq:fedAvg}
\end{equation}

As noted in~\cite{fallah2020personalized} and~\cite{jiang2023improving}, this optimization mechanism is equivalent to Reptile~\cite{nichol2018first}, which samples $\{d_1, d_2, \ldots, d_{n-1}\}$ partitions disjoint to $d_n$ from Omniglot~\cite{lake2015human} or Mini-ImageNet~\cite{vinyals2016matching}, leaving a category out for meta-learning evaluation. Authors then try to few-shot the categories found in the later partition, which are originally contained in $D$.

Note that when $T = 1$ and, for any partition, $k > \left| d_n \right|$, the averaging of model differences is performed just once, but multiple updates on the dataset $d_n$ are conducted. This process creates an ensemble of specialist models that are subsequently merged into a generalist one. In this scenario, authors in~\cite{ilharco2023editing} consider $\tau = U^k(d_n; \theta_0) - \theta_0$ a \textit{task vector}. Additionally, in~\cite{ortizjimenez2023task}, it is highlighted how an overspecialization ($k \gg \left| d_n \right|$) can lead to bad generalization due to lose of ``linearity" of the training regime.

In this work, we initialize $\theta_0$ by training on a Latin dataset $Z$. This ensures that $\theta_0 = U(Z, \theta)$ does not diverge significantly when updating on different domains in Latin examples for handwritten and scene text recognition, which we consider it to be our source dataset $D$. In this way, as recommended by~\cite{ortizjimenez2023task}, we ensure that locality is maintained during the meta-learning process. This is something that Reptile addresses by stating an scenario where the different distributions are sampled by an original one, therefore models do not require to accumulate very different visual features.

Because our scenario should be accountant of the inherent constraints of Document Intelligence (security, sparsity, unbalance...), we advocate for the Federated set-up. This ensures a transferability inherent to Reptile while maintaining the datasets disjoint and, therefore, ensuring privacy. For doing so, it is necessary to start from a reasonable seed in a task-arithmetic manner. 

In contrast to~\cite{nichol2018first}, we utilize meta-learning to adapt to data that strongly differs from the original distribution, this is indicated by a target dataset $g$. While current research focused on evaluating model-merging either in zero-shot or few-shot in-domain examples; we demonstrate how federated learning can enhance reading systems that need to adapt to scenarios such as historical ciphered alphabets or low-resource languages while retaining distributed advantages in a setting where the different partitions $d_n$ originate from disjoint data distributions. In the following sections, we will describe the characteristics of these partitions.

\subsection{Optimization}

This work demonstrates the capabilities of fused models in operating in transfer-learning scenarios. Following the notation used in \cite{ilharco2023editing} and Section~\ref{sec:notation}, we consider a \textit{model} as vector of parameters $\theta \in \R^m$. Given a pretrained model, $\theta_{0}$, we fine-tune it with a training set $D = \{d_1, d_2, \ldots, d_n\}$, which converges towards $\theta^D_T$ (see Figure~\ref{fig:domains_a}), analogously, a sub-domain $d_n \in D$ optimizes models that converge towards $\theta^{d_n}_T$. 

Since a model $\theta^{d_n}$ is optimized for each sub-domain of the original dataset, it represents a mode-specific parametrization, $U^k(d_n, \theta)$. Our first contribution is to empirically demonstrate how the aggregation of such parameters, yields better results on out-of-domain data. In order to perform such aggregation, a \textit{task vector} is produced, corresponding to the vector between the pretrained and the finetuned models $\tau^0_n = U^k(d_n, \theta_{0}) - \theta_{0}$; we then average the respective task vectors of each sub-domain to incorporate it to the pretrained model (see Figure~\ref{fig:domains_b}) such that
\begin{equation}\label{eq:metode}
\theta_{t+1} = \theta_{t} + \frac{1}{N}\sum \tau^{t}_n.    
\end{equation} 
This process, which resembles to \cite{ilharco2023editing} model editing, is repeated $T$ times by averaging the subsequent produced models. As it was previously introduced, the process of iteratively performing task arithmetic is equivalent to training on $T$ Reptile \cite{nichol2018first} steps.
After optimizing a generalized Latin character recognition model for $T$ distributed rounds, with $k$ stochastic optimization steps, we introduce a dataset $g$ that includes samples from an unseen, low-resource alphabet, such as historical ciphers or non-Latin scripts.

Our approach leverages the simplicity of fine-tuning these generalized models, enabling them to outperform models pre-trained in a centralized manner. Moreover, this method does not account for the nature of $g$. Therefore, in contrast to common approaches to low-resource optical character recognition \cite{souibgui2022few}, our method performs with independence of prototypes, proxies or data augmentation.

% \end{table*}

\section{Experimental Setup}
\label{sec:exp}

\begin{figure}
    \centering
     \resizebox{.95\textwidth}{!}{
    \begin{tabular}{ccccc}
        \toprule
        \begin{subfigure}{0.18\textwidth}
            \centering
            \includegraphics[width=\textwidth]{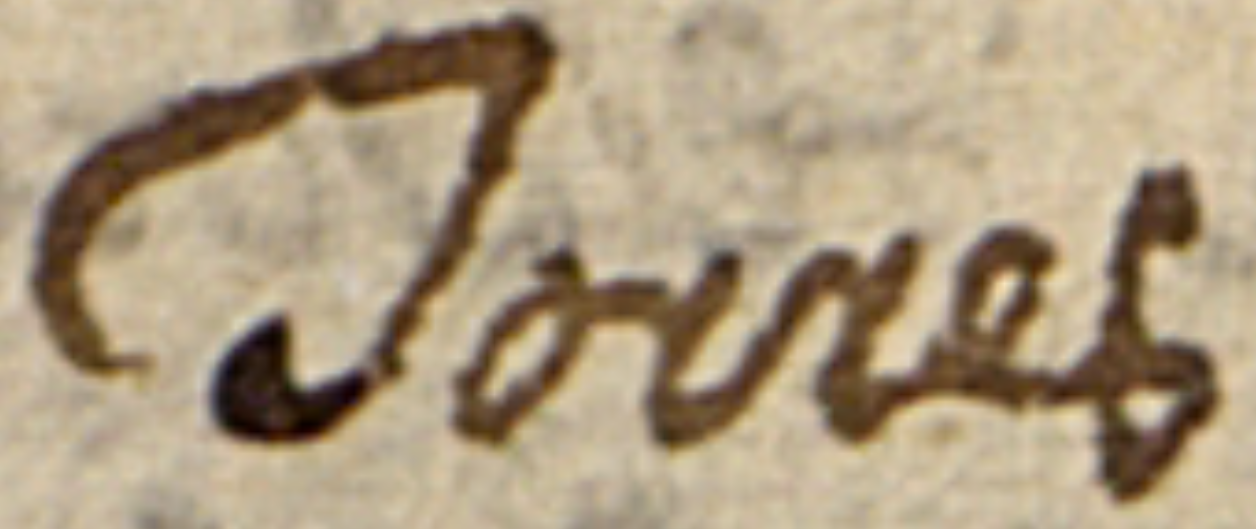}
            
            \caption{ \cite{romero2013esposalles} }
            \label{fig:example_esposalles}
        \end{subfigure} &
        \begin{subfigure}{0.18\textwidth}
            \centering
            \includegraphics[width=\textwidth]{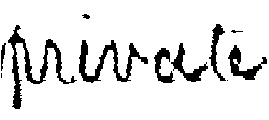}
            \caption{ \cite{fischer2012lexicon}}
            \label{fig:gw}

        \end{subfigure} &
        \begin{subfigure}{0.18\textwidth}
            \centering
            \includegraphics[width=\textwidth]{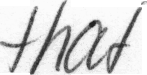}
            \caption{ \cite{marti2002iam}}
                        \label{fig:iam}

        \end{subfigure} &
        \begin{subfigure}{0.18\textwidth}
            \centering
            \includegraphics[width=\textwidth]{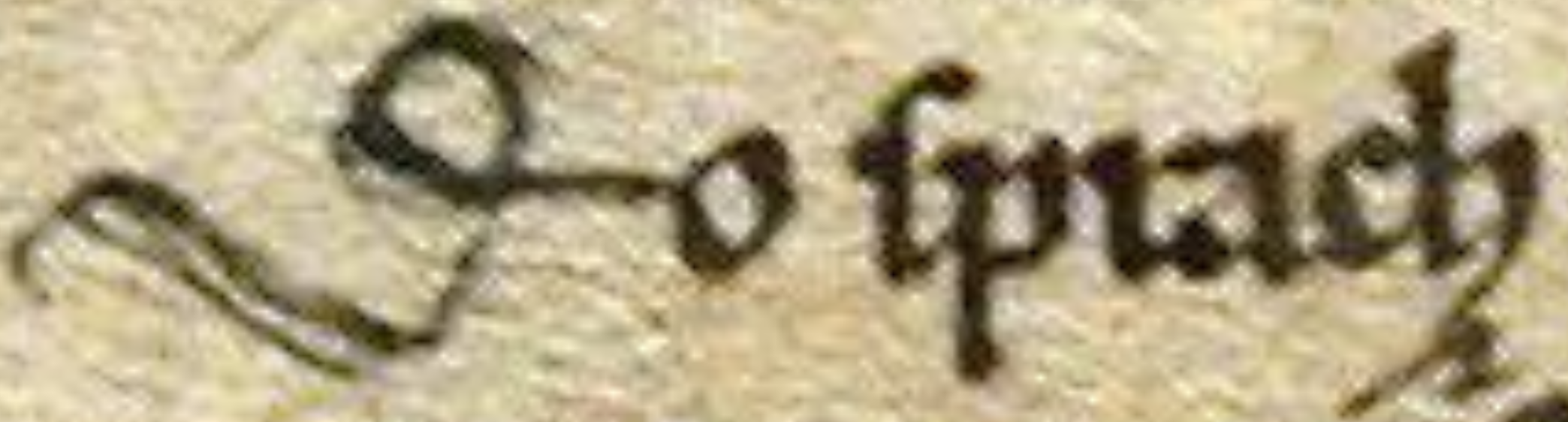}
            \caption{ \cite{fischer2012lexicon}}
                        \label{fig:parzival}

        \end{subfigure} &
        \begin{subfigure}{0.18\textwidth}
            \centering
            \includegraphics[width=\textwidth]{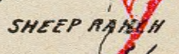}
            \caption{ \cite{weinman2019deep}}
                        \label{fig:maps}

        \end{subfigure} \\
        \midrule
        \begin{subfigure}{0.18\textwidth}
            \centering
            \includegraphics[width=\textwidth]{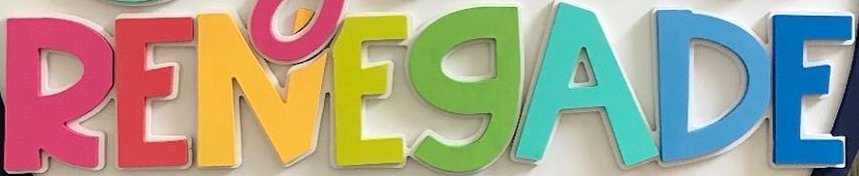}
            \caption{From WA\footnotemark{}}
            \label{fig:wa}
        \end{subfigure} &
        \begin{subfigure}{0.18\textwidth}
            \centering
            \includegraphics[width=\textwidth]{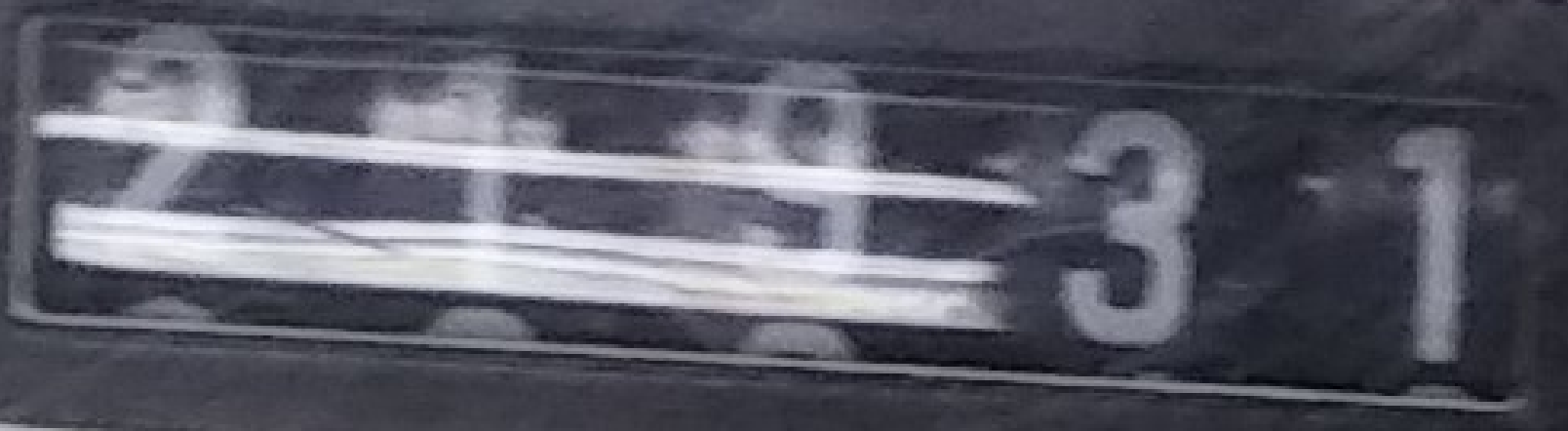}
            \caption{ \cite{laroca2019convolutional}}
            \label{fig:amr}
        \end{subfigure} &
        \begin{subfigure}{0.18\textwidth}
            \centering
            \includegraphics[width=\textwidth]{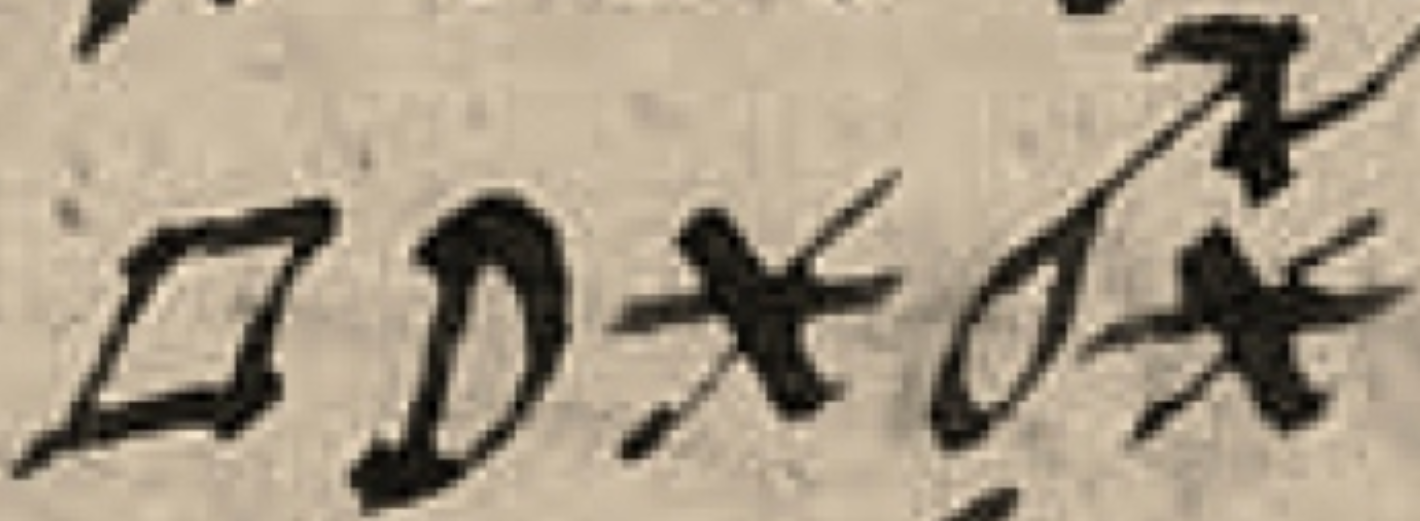}
            \caption{ \cite{aldarrabBorgLat8982018}}
            \label{fig:borg}
        \end{subfigure} &
        \begin{subfigure}{0.18\textwidth}
            \centering
            \includegraphics[width=\textwidth]{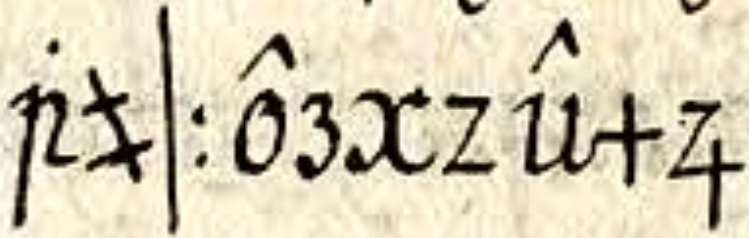}
            \caption{ \cite{knightCopialeCipher2011}}
            \label{fig:copiale}
        \end{subfigure} &
        \begin{subfigure}{0.18\textwidth}
            \centering
            \includegraphics[width=\textwidth]{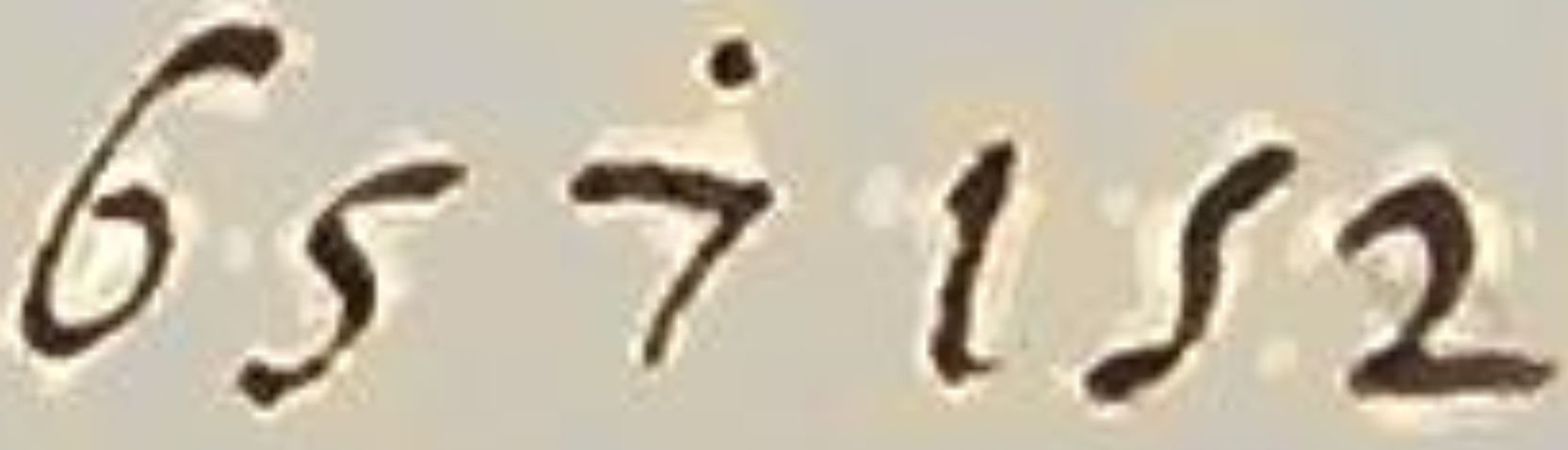}
            \caption{ \cite{heder2022decode}}
                \label{fig:vatican}

        \end{subfigure} \\
        \midrule
        \begin{subfigure}{0.18\textwidth}
            \centering
            \includegraphics[width=\textwidth]{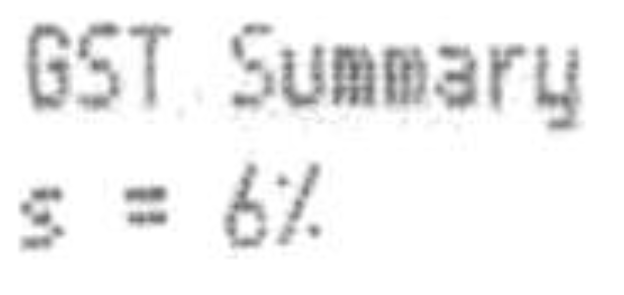}
            \caption{ \cite{karatzas2013icdar}}
                        \label{fig:sroie}

        \end{subfigure} &
        \begin{subfigure}{0.18\textwidth}
            \centering
            \includegraphics[width=\textwidth]{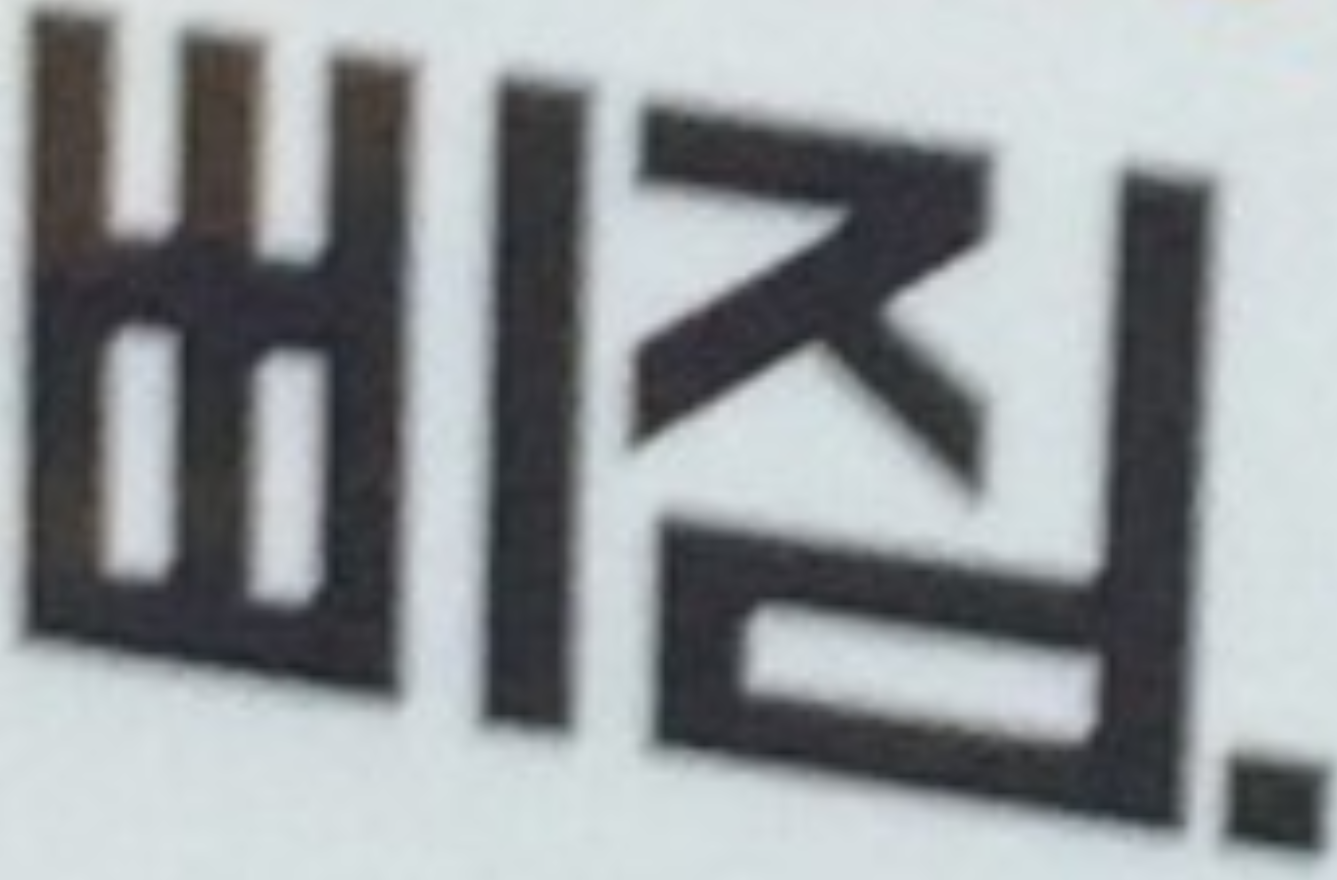}
            \caption{Korean-\cite{nayef2019icdar2019}}
                        \label{fig:korean}

        \end{subfigure} &
        \begin{subfigure}{0.18\textwidth}
            \centering
            \includegraphics[width=\textwidth]{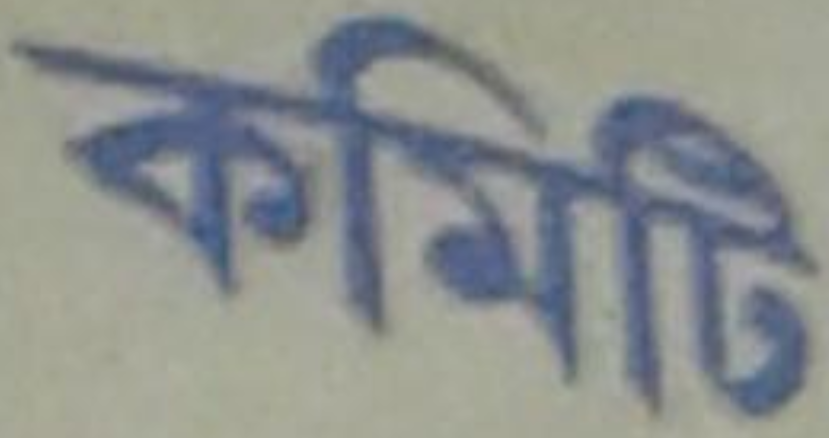}
            \caption{Bangla-\cite{nayef2019icdar2019}}
                        \label{fig:bangla}

        \end{subfigure} &
        \begin{subfigure}{0.18\textwidth}
            \centering
            \includegraphics[width=\textwidth]{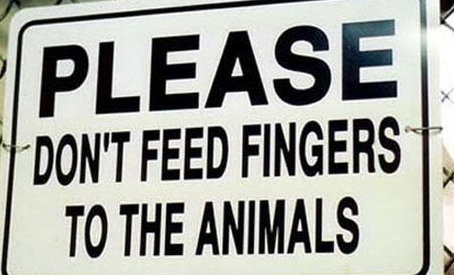}
            \caption{Latin-\cite{nayef2019icdar2019}}
                                    \label{fig:mlt19}

        \end{subfigure} &
        \begin{subfigure}{0.18\textwidth}
            \centering
            \includegraphics[width=\textwidth]{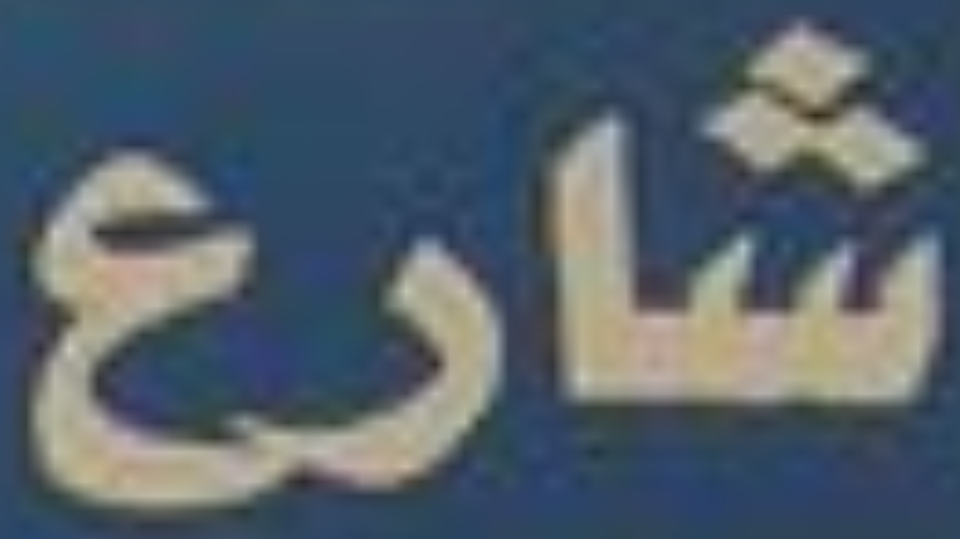}
            \caption{Arabic-\cite{nayef2019icdar2019}}
                                                \label{fig:arabic}

        \end{subfigure} \\
        \midrule
        \begin{subfigure}{0.18\textwidth}
            \centering
            \includegraphics[width=\textwidth]{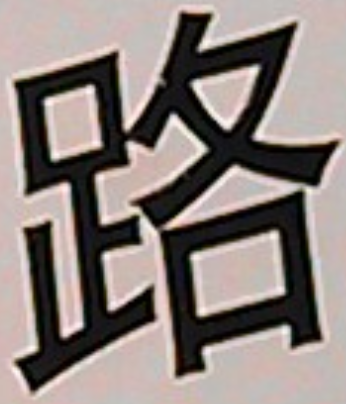}
            \caption{Chinese-\cite{nayef2019icdar2019}}                                    \label{fig:chinese}

        \end{subfigure} &
        \begin{subfigure}{0.18\textwidth}
            \centering
            \includegraphics[width=\textwidth]{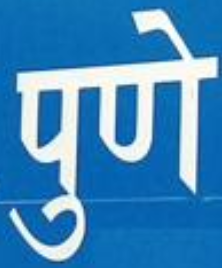}
            \caption{Hindi-\cite{nayef2019icdar2019}}
                                                \label{fig:hindi}

        \end{subfigure} &
        \begin{subfigure}{0.18\textwidth}
            \centering
            \includegraphics[width=\textwidth]{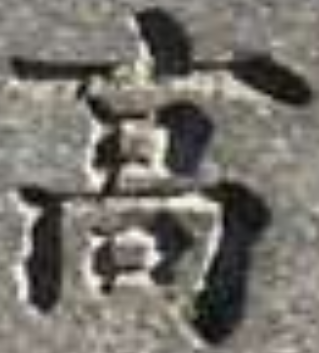}
            \caption{Japanese-\cite{nayef2019icdar2019}}
                                                \label{fig:japanese}

        \end{subfigure} &
        \begin{subfigure}{0.18\textwidth}
            \centering
            \includegraphics[width=\textwidth]{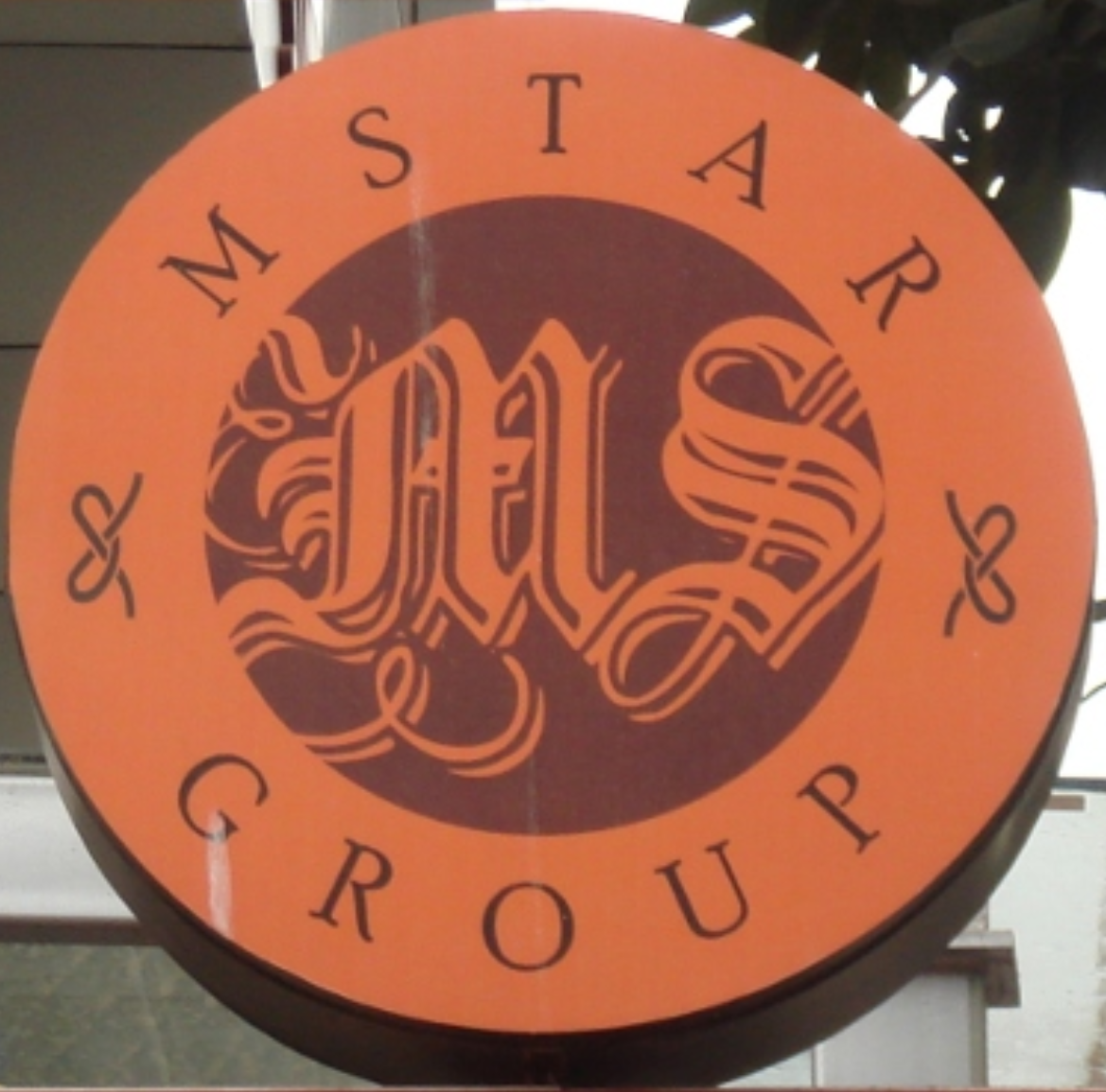}
            \caption{ \cite{ch2020total}}                                    \label{fig:total}

        \end{subfigure} &
        \begin{subfigure}{0.18\textwidth}
            \centering
            \includegraphics[width=\textwidth]{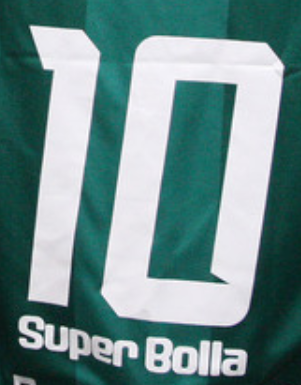}
            \caption{ \cite{singh2021textocr}}
                                                \label{fig:textocr}

        \end{subfigure} \\
        \bottomrule
    \end{tabular}}
    \caption{Landscape of some of the datasets used in this research: Esposalles (a), George Washington (b), IAM (c), Parzival (d), Historical Maps (e), WordArt (f), AMR (g), Borg Cipher (h), Copiale Cipher (i), Vatican Cipher (j), SROIE (k), MLT (l to r), TotalText (s), and TextOCR (t).}

    \label{fig:all_datasets}
\end{figure}
\footnotetext{\url{codalab.lisn.upsaclay.fr/competitions/17182}}

In this section, we expose the experimental set up used to develop the ideas previously introduced, primarily, that merging expert models, drastically boosts the incorporation of low-resource and historical data. Therefore, we introduce a common architecture for all models to be optimized, among its technical details. We also provide details about the used data and domains (in-domain, out-of-domain and new scripts). Finally, a baseline is proposed to perform a non-biased comparison with a common hyperparameterization and architecture.

\subsection{Datasets}

\label{sec:data}
As it was previously introduced, this research takes the assumption of the existence of a dataset $D$ composed by sub-domains $\{d_1, \ldots, d_n\}$, a group of testing samples out-of-distribution $x \notin D$ and a target new domain $g$. 

Since the experiments are conducted through the context of intelligent reading systems, in this section we introduce the hierarchy of datasets which this research will utilize. First, as it was previously introduced, HierText\cite{long2022towards} ($Z$) comprises around 11K Latin-script documents and natural scenes. This dataset is used in our experiments to produce an initial pretrained model $\theta_{0}$. In Figure~\ref{fig:all_datasets}, some samples from the used datasets are presented. In this section, we discuss about how this data is arranged to construct this different sets of in-domain, out-of-domain and transfer learning tasks.
\begin{figure}[t]
    \centering
    \includegraphics[width=0.93\linewidth]{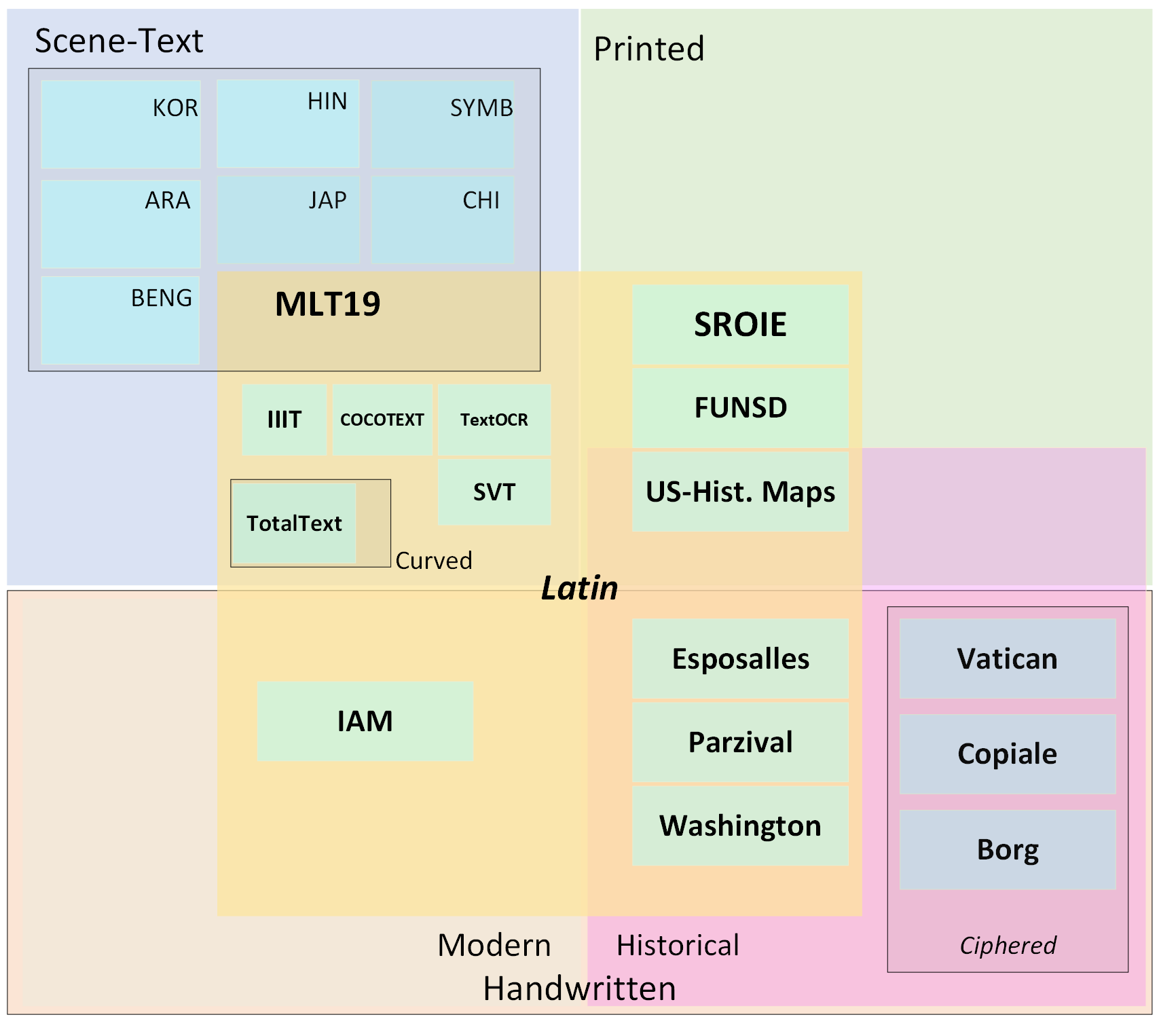}
    \caption{Taxonomy (Venn diagram) of the datasets used in this work. We categorize data into scene, printed, and handwritten domains while accounting for distribution shifts in historical documents. Additionally, datasets are classified based on their alphabet type: Latin, ciphered, and non-Latin.}
    \label{fig:jerarquia}
\end{figure}
For in-domain training, we primarily use the domain of handwritten (\textbf{HW}) and scene text (\textbf{Scene}) recognition, on which we list Esposalles \cite{romero2013esposalles}, IAM \cite{marti2002iam}, George Washington \cite{fischer2012lexicon} and Parzival (Figures \ref{fig:example_esposalles} to \ref{fig:parzival}) for handwritten and CoCoText \cite{veit2016cocotext} and TextOCR \cite{singh2021textocr} for scenes. Additionally, MLT19 \cite{nayef2019icdar2019} is considered as training set of the scene domain in its Latin subsampling; while the rest of splits (Arabic, Chinese, Japanese, Korean, Hindi and Bangla) in MLT19 are kept as fine-tuning data for transfer learning experiments (see Figures \ref{fig:arabic}, \ref{fig:chinese}, \ref{fig:japanese}, \ref{fig:korean}, \ref{fig:hindi}, \ref{fig:bangla}). SVT \cite{wang2011end} and IIIT5K \cite{mishra2012scene} are kept as evaluation data for scene text recognition as common practice in the literature \cite{chen2021text}.

Other data considered to experiment with the incorporation of new scripts are ciphered historical documents: Vatican \cite{heder2022decode}, Copiale \cite{knightCopialeCipher2011} and Borg \cite{aldarrabBorgLat8982018} (see Figures \ref{fig:vatican}, \ref{fig:copiale}, \ref{fig:borg}); which constitute the biggest challenge since it contains unseen handwritten symbols (with the exception of Vatican, containing a mixture of numbers, symbols and special accents) without further contextual information, such as background or incidental correlations, to rely on. 

Importantly, we define a set which we consider ``out-of-domain'' for different reasons. Starting with printed documents, historical maps \cite{weinman2019deep} are used as samples containing information not strictly related with text (see Figure \ref{fig:maps}). Other kind of printed documents, such as receipts in SROIE \cite{karatzas2013icdar} or forms in FUNSD \cite{jaume2019funsd} face challenges associated with arbitrary segments of text (such as named entities, codes or numbers), which, as noted by \cite{wan2020vocabulary}, reduces the reliance on shortcuts with respect data in $D$, on which one can associate words to real-world symbols or to build an implicit language model, which is the case for \cite{romero2013esposalles} with sentences having a fixed structure. Similarly, the AMR dataset \cite{laroca2019convolutional} contains digits from domestic gas company meters, which contains arbitrary sequences combined with noise and flares (see Figure \ref{fig:amr}). Lastly, TotalText\cite{ch2020total} and WordArt datasets (see Figure \ref{fig:wa}), contain curved texts with artistic fonts. We consider this an important domain shift because ViTs are trained upon the assumption of vertically aligned text; on which the positional encoding plays a vital role. 

For most of the examples shown, we use the original training / test splits, which are publicly available at\footnote{Contact main corresponding author} among the partitions randomly selected in some cases such as Esposalles, WA or MLT19. In the case of IAM dataset, we made use of widely considered \cite{michael2019evaluating,singh2021full,kang2019convolve} AAChen splits.

Practitioners can find a visual summary of the defined taxonomy in Figure~\ref{fig:jerarquia}. In short: we use the datasets IAM, Esposalles, Parzival and GW, as well as $\text{MLT}_{Latin}$, CoCoText, and TextOCR for training. Additionally, we evaluate scene text on SVT and IIIT. New languages and scripts are separately introduced through the datasets $\text{MLT}_{Arabic}$, $\text{MLT}_{Korean}$, $\text{MLT}_{Bangla}$, $\text{MLT}_{Chinese}$, $\text{MLT}_{Hindi}$, and $\text{MLT}_{Japanese}$, as well as Copiale, Borg, and Vatican. HierText is used for setting and initial seed ``zero-shot'' ($Z$) and the rest of datasets are kept for evaluating out-of-domain performance.

\subsection{Architecture Details}
To ensure proper comparisons, we establish a common baseline across various in-domain training, evaluation and transfer learning splits. We utilize a ViT encoder as described in \cite{atienza2021vision}. However, we enhance the performance of the model by incorporating an additional encoding layer on top of the previous output, as suggested in \cite{etter2023hybrid}. This additional layer includes a positional encoding and is conditioned to a projection of the non-positionally-encoded input.
\begin{table*}[t]

            \caption{Reported baseline accuracy on the different handwritten, scene and out-of-distribution datasets by fine-tunning a pretrained ViT-CTC model with HierText \cite{long2022towards}. }
    \label{tab:ft_from_hiertext_baseline}
    \resizebox{\textwidth}{!}{
    \begin{tabular}{lrrrrrrrrrrrrrrr}
\hline
\multicolumn{16}{c}{Accuracy on Finetuned ViT Models (from HierText)} \\ \hline
\multicolumn{1}{c|}{\multirow{3}{*}{Train Set}} &
  \multicolumn{15}{c}{Test Set} \\ \cline{2-16} 
\multicolumn{1}{c|}{} &
  \multicolumn{4}{c|}{HW} &
  \multicolumn{5}{c|}{Scene} &
  \multicolumn{6}{c}{OOD} \\ \cline{2-16} 
\multicolumn{1}{c|}{} &
  \multicolumn{1}{c}{IAM} &
  \multicolumn{1}{c}{Esposalles} &
  \multicolumn{1}{c}{Parzival} &
  \multicolumn{1}{c|}{GW} &
  \multicolumn{1}{c}{MLT} &
  \multicolumn{1}{c}{CoCo} &
  \multicolumn{1}{c|}{TOCR} &
  \multicolumn{1}{c}{SVT} &
  \multicolumn{1}{c|}{IIIT} &
  \multicolumn{1}{c}{WA} &
  \multicolumn{1}{c}{FUNSD} &
  \multicolumn{1}{c}{HM} &
  \multicolumn{1}{c}{TT} &
  \multicolumn{1}{c}{Sroie} &
  \multicolumn{1}{c}{AMR} \\ \midrule
\multicolumn{1}{l|}{IAM} &
  \textbf{.798} &
  .312 &
  .166 &
  \multicolumn{1}{r|}{.284} &
  .682 &
  .555 &
  \multicolumn{1}{r|}{.608} &
  .783 &
  \multicolumn{1}{r|}{.825} &
  .567 &
  .784 &
  .743 &
  .581 &
  .450 &
  .645 \\
\multicolumn{1}{l|}{Esposalles} &
  .462 &
  \textbf{.980} &
  .186 &
  \multicolumn{1}{r|}{.235} &
  .625 &
  .547 &
  \multicolumn{1}{r|}{.568} &
  .803 &
  \multicolumn{1}{r|}{.798} &
  .487 &
  .679 &
  .683 &
  .617 &
  .380 &
  .557 \\
\multicolumn{1}{l|}{Parzival} &
  .520 &
  .371 &
  .884 &
  \multicolumn{1}{r|}{.236} &
  .661 &
  .557 &
  \multicolumn{1}{r|}{.610} &
  .789 &
  \multicolumn{1}{r|}{.803} &
  .527 &
  .754 &
  .726 &
  .577 &
  .427 &
  .620 \\
\multicolumn{1}{l|}{GW} &
  .638 &
  .377 &
  .208 &
  \multicolumn{1}{r|}{.774} &
  .700 &
  .576 &
  \multicolumn{1}{r|}{.636} &
  .808 &
  \multicolumn{1}{r|}{.843} &
  .593 &
  .775 &
  .759 &
  .599 &
  .448 &
  .655 \\ \midrule
\multicolumn{1}{l|}{HW} &
  .782 &
  .978 &
  \textbf{.898} &
  \multicolumn{1}{r|}{\textbf{.779}} &
  .624 &
  .529 &
  \multicolumn{1}{r|}{.547} &
  .817 &
  \multicolumn{1}{r|}{.793} &
  .524 &
  .701 &
  .684 &
  .585 &
  .363 &
  .631 \\ \midrule
\multicolumn{1}{l|}{MLT} &
  .443 &
  .315 &
  .137 &
  \multicolumn{1}{r|}{.108} &
  \textbf{.800} &
  .511 &
  \multicolumn{1}{r|}{.565} &
  .766 &
  \multicolumn{1}{r|}{.801} &
  .528 &
  .749 &
  .746 &
  .561 &
  .440 &
  .603 \\
\multicolumn{1}{l|}{CoCo} &
  .563 &
  .337 &
  .181 &
  \multicolumn{1}{r|}{.251} &
  .660 &
  .598 &
  \multicolumn{1}{r|}{.598} &
  .842 &
  \multicolumn{1}{r|}{\textbf{.852}} &
  .575 &
  .726 &
  .756 &
  .613 &
  \textbf{.497} &
  .642 \\
\multicolumn{1}{l|}{TOCR} &
  .622 &
  .416 &
  .236 &
  \multicolumn{1}{r|}{.366} &
  .719 &
  \textbf{.620} &
  \multicolumn{1}{r|}{\textbf{.768}} &
  \textbf{.896} &
  \multicolumn{1}{r|}{.843} &
  \textbf{.652} &
  .813 &
  \textbf{.842} &
  \textbf{.679} &
  .471 &
  \textbf{.710} \\ \midrule
\multicolumn{1}{l|}{Scene} &
  .511 &
  .274 &
  .194 &
  \multicolumn{1}{r|}{.220} &
  .767 &
  .588 &
  \multicolumn{1}{r|}{.712} &
  .851 &
  \multicolumn{1}{r|}{.829} &
  .561 &
  .785 &
  .796 &
  .632 &
  .459 &
  .685 \\ \midrule
\multicolumn{1}{l|}{All Above} &
  .773 &
  .963 &
  .858 &
  \multicolumn{1}{r|}{.753} &
  .742 &
  .592 &
  \multicolumn{1}{r|}{.722} &
  .863 &
  \multicolumn{1}{r|}{.833} &
  .585 &
  .791 &
  .806 &
  .640 &
  .473 &
  .706 \\ \midrule
\multicolumn{1}{l|}{Zero-Shot} &
  .658 &
  .395 &
  .249 &
  \multicolumn{1}{r|}{.389} &
  .714 &
  .585 &
  \multicolumn{1}{r|}{.664} &
  .826 &
  \multicolumn{1}{r|}{.807} &
  .627 &
  \textbf{.814} &
  .792 &
  .612 &
  .465 &
  .685 \\ \bottomrule
\end{tabular}

    }

\end{table*}
Building upon previous literature \cite{campiotti2022optical,etter2023hybrid,sahu2015sequence}, we train the transformer using a CTC \cite{graves2006connectionist} loss function. Unless stated otherwise, it is assumed to use 75 epochs of optimizing with Adam, a learning rate of $10^{-5}$ and 128 batch size. In the context of learning new alphabets, 30 epochs are used instead in order to promote models that can easily incorporate new knowledge; this hyperparametrization is empirically defined to maintain a common regime to all models while ensuring convergence in all cases.

As pointed out in \cite{ortizjimenez2023task}, task arithmetic is closely related to the idea of locality during optimization, meaning that the results are not reproducible in scenarios where models diverge by a considerable distance. As noted by the original authors \cite{ilharco2023editing}, an initial reasonably good seed is needed (pretraining stage). In our study, we established the seed through 75 epochs of pre-training on recognition utilizing the HierText dataset \cite{long2022towards}. This dataset comprises 11,639 Latin script images extracted from natural scenes and documents. During the pre-training stage, the model was trained to recognize 749,802 words across various scenarios, ensuring its centrality and capability to perform task arithmetic effectively.

\subsection{Baselines}
\label{sec:bas}
The literature in optical character recognition (OCR) has exhibited a sparse trend over the years, where handwritten tasks are often intertwined with word spotting (retrieval) and recognition tasks \cite{nikolaidou2022survey}. In text recognition, authors typically present results by comparing different methods with available evaluation metrics specific to the dataset \cite{chen2021text}. Some platforms \cite{robust_reading_competition}, have achieved considerable success in standardizing reading systems based on a common baseline. However, to the best of our knowledge, there is a lack of joint analysis of word-recognition models in the literature. Similar to how certain fields like metric learning have established common bases \cite{musgrave2020metric}, we propose a fair evaluation of the different domains with fixed model and hyperparameters.

The resulting accuracy on both in-domain (handwritten and scene) and out-of-domain test sets is reported in Table~\ref{tab:ft_from_hiertext_baseline}. Additionally, supplementary material provides an extensive range of additional results encompassing the rest of the metrics. The baseline indeed reveals some expected outcomes. The initial model may outperform others in contexts like WA, FUNSD, and SROIE, where documents are prevalent in the HierText dataset. However, for other examples, training with additional data slightly improves generalization. With this baseline established, our study is well-positioned to further widen this gap towards better generalization.

\section{Results and Discussion}
\label{sec:res}
In the following section, results are presented for both generalization and transfer learning. In order to build models capable of quickly incorporating new knowledge from data, it is fundamental to build a model which is generic enough to incorporate such behavior. Therefore, we first evaluate the introduced strategies on generalization. Later on, we use the best generalizing model as a seed for introducing novel scripts to the recognition system and quantitative compare how well adapted those models are with respect to the baseline (training from $\theta_{0}$).
\begin{table*}[t]

    \caption{Comparison on the accuracy of previous best models according to the baseline (FT-Best), the model comprising a fine-tune with all the previous datasets (Handwritten, Scene) and models resulting from averaging task vectors (Avg). Note that \texttt{FT-Best} is not a single model, rather the best baseline: it acts as an {\ul upper bound} for total performance. Best result for models is \textbf{marked as bold font}.}
    \centering
    \resizebox{\textwidth}{!}{\begin{tabular}{lrrrrrrrrrrrrrrr}
\toprule
\multicolumn{16}{c}{Accuracy on ViT Models (from HierText)}                                                                                                                                                                                                                                                                                                                                                                                                                                               \\ \midrule
\multicolumn{1}{c|}{\multirow{3}{*}{Approach}} & \multicolumn{15}{c}{Test Set}                                                                                                                                                                                                                                                                                                                                                                                                                            \\ \cline{2-16} 
\multicolumn{1}{c|}{}                          & \multicolumn{4}{c|}{HW}                                                                                                      & \multicolumn{5}{c|}{Scene}                                                                                                                                   & \multicolumn{6}{c}{OOD}                                                                                                                                    \\ \cline{2-16} 
\multicolumn{1}{c|}{}                          & \multicolumn{1}{c}{IAM} & \multicolumn{1}{c}{Esposalles} & \multicolumn{1}{c}{Parzival} & \multicolumn{1}{c|}{GW}            & \multicolumn{1}{c}{MLT} & \multicolumn{1}{c}{CoCo} & \multicolumn{1}{c|}{TOCR}          & \multicolumn{1}{c}{SVT} & \multicolumn{1}{c|}{IIIT}                & \multicolumn{1}{c}{WA} & \multicolumn{1}{c}{FUNSD} & \multicolumn{1}{c}{HM} & \multicolumn{1}{c}{TT} & \multicolumn{1}{c}{Sroie} & \multicolumn{1}{c}{AMR} \\ \midrule
\multicolumn{1}{l|}{FT-Best}                   & {\ul .798}              & {\ul .980}                     & {\ul .884}                   & \multicolumn{1}{r|}{{\ul .779}}    & {\ul .800}              & .620                     & \multicolumn{1}{r|}{{\ul .768}}    & {\ul .896}              & \multicolumn{1}{r|}{.843}                & .652                   & .814                      & {\ul .842}             & {\ul .679}             & .473                      & .710                    \\
\multicolumn{1}{l|}{FT (HW, Scene)}            & \textbf{.773}           & \textbf{.963}                  & \textbf{.858}                & \multicolumn{1}{r|}{\textbf{.753}} & \textbf{.742}           & .592                     & \multicolumn{1}{r|}{.722}          & .863                    & \multicolumn{1}{r|}{.833}                & .585                   & .791                      & .806                   & .640                   & .473                      & .706                    \\ \midrule
\multicolumn{1}{l|}{Avg (HW, Scene) (ours)}           & .747                    & .808                           & .567                         & \multicolumn{1}{r|}{.638}          & .729                    & .623                     & \multicolumn{1}{r|}{\textbf{.728}} & \textbf{.891}           & \multicolumn{1}{r|}{.847}                & .648                   & .821                      & .813                   & .674                   & .474                      & .738                    \\
\multicolumn{1}{l|}{Avg (Ind.) (ours)}                & .708                    & .606                           & .412                         & \multicolumn{1}{r|}{.523}          & .730                    & .618                     & \multicolumn{1}{r|}{.713}          & .873                    & \multicolumn{1}{r|}{.846}                & .675                   & {\ul \textbf{.835}}       & .824                   & .663                   & .483                      & .715                    \\
\multicolumn{1}{l|}{Avg (Orth.) (ours)}               & .736                    & .696                           & .303                         & \multicolumn{1}{r|}{.565}          & .727                    & {\ul \textbf{.627}}      & \multicolumn{1}{r|}{.721}          & .880                    & \multicolumn{1}{r|}{{\ul \textbf{.855}}} & {\ul \textbf{.684}}    & {\ul \textbf{.835}}       & \textbf{.829}          & \textbf{.676}          & {\ul \textbf{.484}}       & {\ul \textbf{.733}}     \\ \midrule
\multicolumn{1}{l|}{Ensemble \cite{wick2021one} (Ind.)}           & .513                    & .451                           & .140                         & \multicolumn{1}{r|}{.267}          & .561                    & .559                     & \multicolumn{1}{r|}{.513}          & .666                    & \multicolumn{1}{r|}{.678}                & .267                     & .564                        & .669                   & .514                   & .286                      & .485                    \\ \bottomrule
\end{tabular}
}

    \label{tab:ood}
\end{table*}

\begin{table}[t]
\caption{
Accuracy results for transfer learning across multiple languages using different strategies. We compare centralized fine-tuning and task arithmetic (distributed), both starting from a shared pretrained model ($Z$). We also evaluate task arithmetic from a randomly initialized model. Despite using the same data splits and number of examples, the distributed strategy consistently outperforms centralized fine-tuning when transferring to low-resource alphabetic systems.
}
\resizebox{\textwidth}{!}{
\begin{tabular}{l|ccc|cccccc|c}
\toprule
 & Vatican & Borg & Copiale & Arabic & Chinese & Japanese & Korean & Bangla & Hindi & $\times\Delta$ \\
\midrule
Baseline U ( Z ) & .549 & .382 & .825 & .131 & .020 & .116 & .282 & .260 & .470 & - \\
Centralized (from Z ) & .465 & .000 & .794 & .175 & .0103 & .0951 & .214 & .115 & .384 & 0.69 \\
Distributed (from Z) & \textbf{.591} & \textbf{.505} & .838 & \textbf{.422} & \textbf{.073} & \textbf{.197} & \textbf{.374} & \textbf{.462} & \textbf{.514} & \textbf{1.62} \\
Distributed (random) & .480 & .272 & \textbf{.930} & .410 & .010 & .114 & .271 & .266 & .282 & 1.04 \\
\bottomrule
  \end{tabular}}

\label{tab:results_lang}
\end{table}

\subsection{Quantitative Analysis}
As it was previously introduced, this research is primarily seeking generalist models as candidates to achieving superior transfer learning capabilities. In Table~\ref{tab:ood}, we compare the following approaches:

\textbf{FT-Best}: Stands for the best result in Table~\ref{tab:ft_from_hiertext_baseline}, which corresponds to the established baseline. This register is composed from different models rather than a single one, therefore the maximum is noted by underscoring. The maximum performance achieved by a single model is noted in \textbf{bold}.

\textbf{FT (HW, Scene)}: Is a finetuned model from $\theta_{0}$ using both scene and handwritten training datasets jointly. 

\textbf{Avg (HW, Scene)}: In this approach, two models are independently trained on handwritten and scene datasets ($d^{hw}, d^{scene}$). The resulting task vectors are then applied to the zero-shot model via averaging. 
\textbf{Avg (Ind.)}: In this case, task vectors are obtaining after dis-jointly training a different model for each training dataset ($\{d_1, \ldots, d_7\}$). We then apply those vectors by averaging them to obtain the final model, which results in $\theta^{ind}$.

\textbf{Avg (Orth.)}: As noted by \cite{ilharco2023editing}, most of the forgetting introduced by task arithmetic principles is due to non-orthogonality of models, in such scenario, \cite{ortizjimenez2023task} attributes the concept of weights being entangled, which can be understood as a summing of destructive signals. In our case, we noticed that models trained on MLT19 and Parzival exhibited non-orthogonality via cosine distance calculation. Therefore,  $\theta^ {parzival}$ and $\theta^{mlt}$ are left out of the merging procedure.

Inspection of Table~\ref{tab:ood} reveals that merging models slightly improves performance across additional domains. This suggests enhanced generalization capabilities, indicating that the resulting models are more flexible in incorporating new knowledge. Compelling evidence of this behavior is presented in the following paragraphs, where we take advantage on this generalizing models to incorporate unseen alphabets. Further and extensive ablation study is to be found later in the manuscript.

Regarding the incorporation of new languages, after performing model merging, we analyze the impact of using  $\theta^{ind}$ as an initialization point for learning previously unseen scripts in a new training stage. As shown in Table~\ref{tab:results_lang}, fine-tuning from the averaged models significantly accelerates the incorporation of new languages, in some cases achieving nearly three times the improvement compared to the baseline. In contrast, models trained directly on in-domain data within a centralized framework rarely outperform the baseline. Even when slight improvements occur, our distributed strategy consistently yields superior results, even when utilizing the same training data.

\begin{figure*}[t]
    \centering
    \begin{subfigure}{0.49\textwidth}
        \centering
        \includegraphics[width=\textwidth]{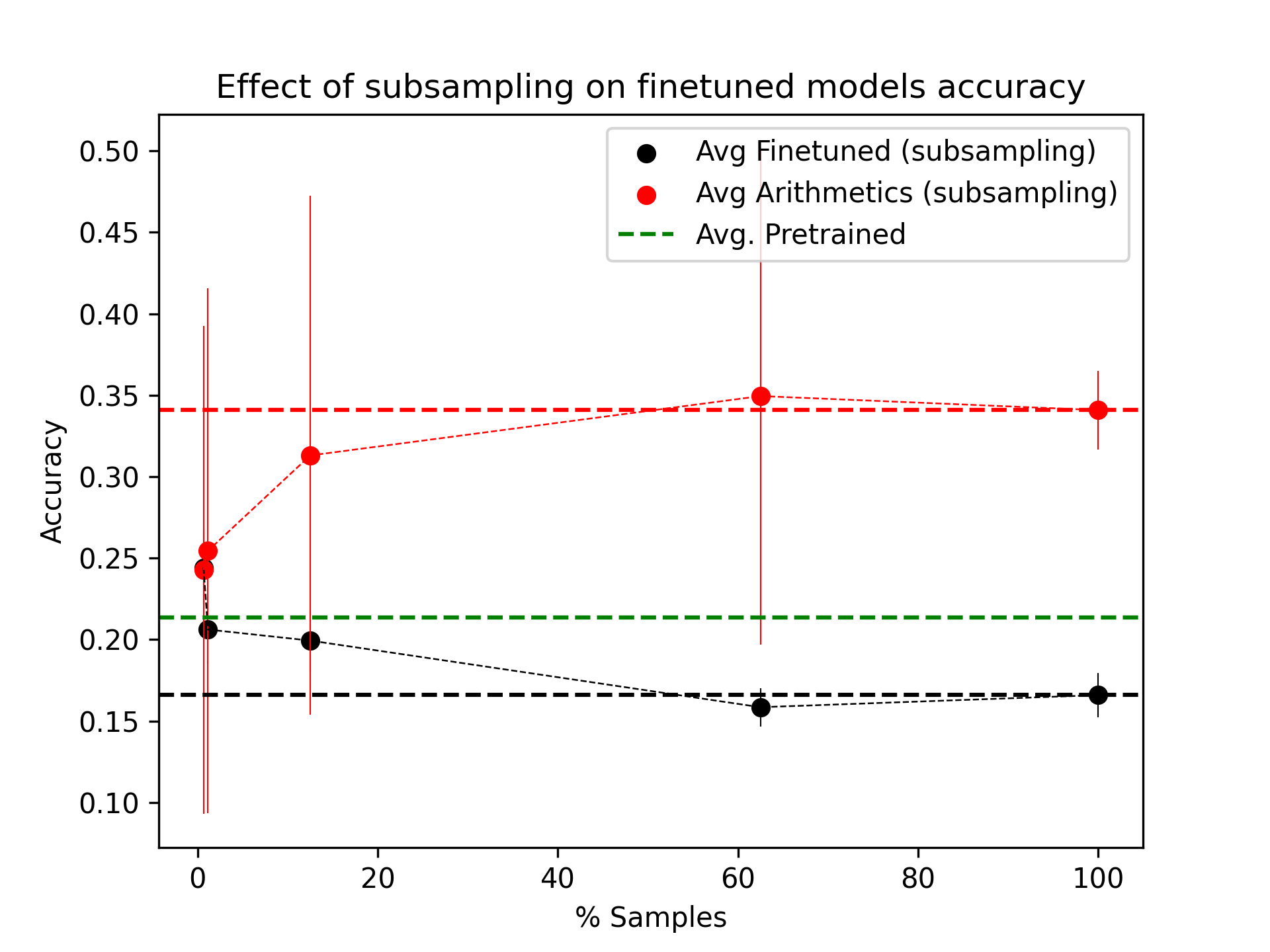}
        \caption{}
        \label{fig:scripts_subsampling}
    \end{subfigure}
        \begin{subfigure}{0.49\textwidth}
        \centering
        \includegraphics[width=\textwidth]{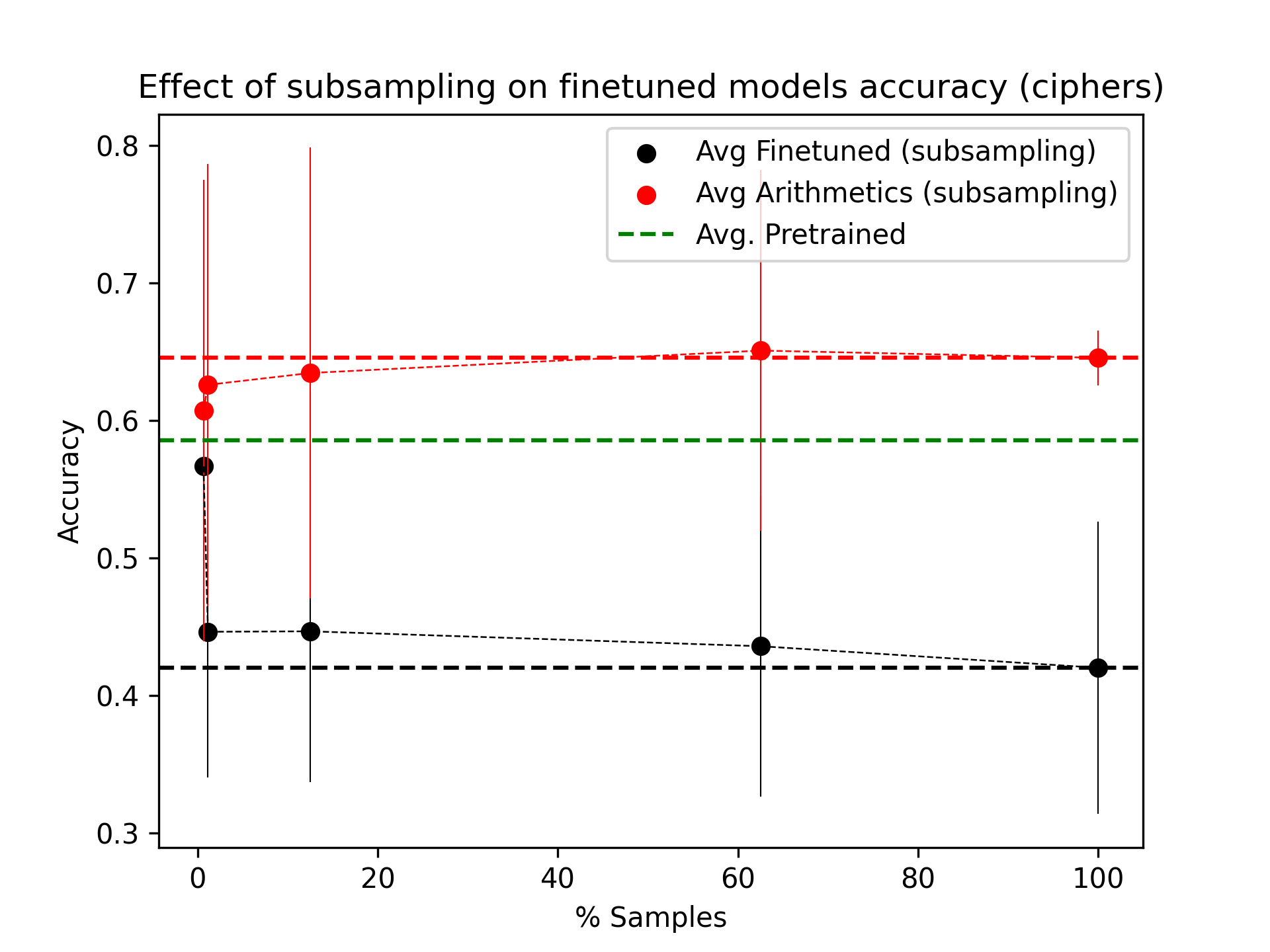}
        \caption{}
        \label{fig:cipher_subsampling}
    \end{subfigure}
    \caption{Average accuracy achieved by our approach (red), baseline (green) and traditional finetuning (black) for new languages (a) and ciphered texts (b) under the same subsampling percentage of pretraining data.}
    \label{fig:scale_actor}
\end{figure*}

Moreover, in Table~\ref{tab:results_lang}, we valiate our design choice of using $U(Z, \theta)$ as seed for our experiments. The reason for the necessity of using a reasonably good seed for performing generalization, rather than a random seed could be attributed to the notion of locality observed in \cite{ilharco2023editing}. This is, randomly initialized models tend to significantly diverge when training on distributed data. Even though our strategy dominates distributed learning from a random seed; the last, tends also to outperform centralized approaches, which proves to be the least practical approach.

However, this scenario assumes that $\theta^{ind}_1$ is a converged model across all datasets after performing task arithmetic. This raises questions regarding both the data used and the decision to use task arithmetic ($T=1$) instead of a more extended meta-learning procedure.

\begin{figure}[t]
    \centering
    \begin{subfigure}{0.32\textwidth}
        \centering
        \includegraphics[width=\textwidth]{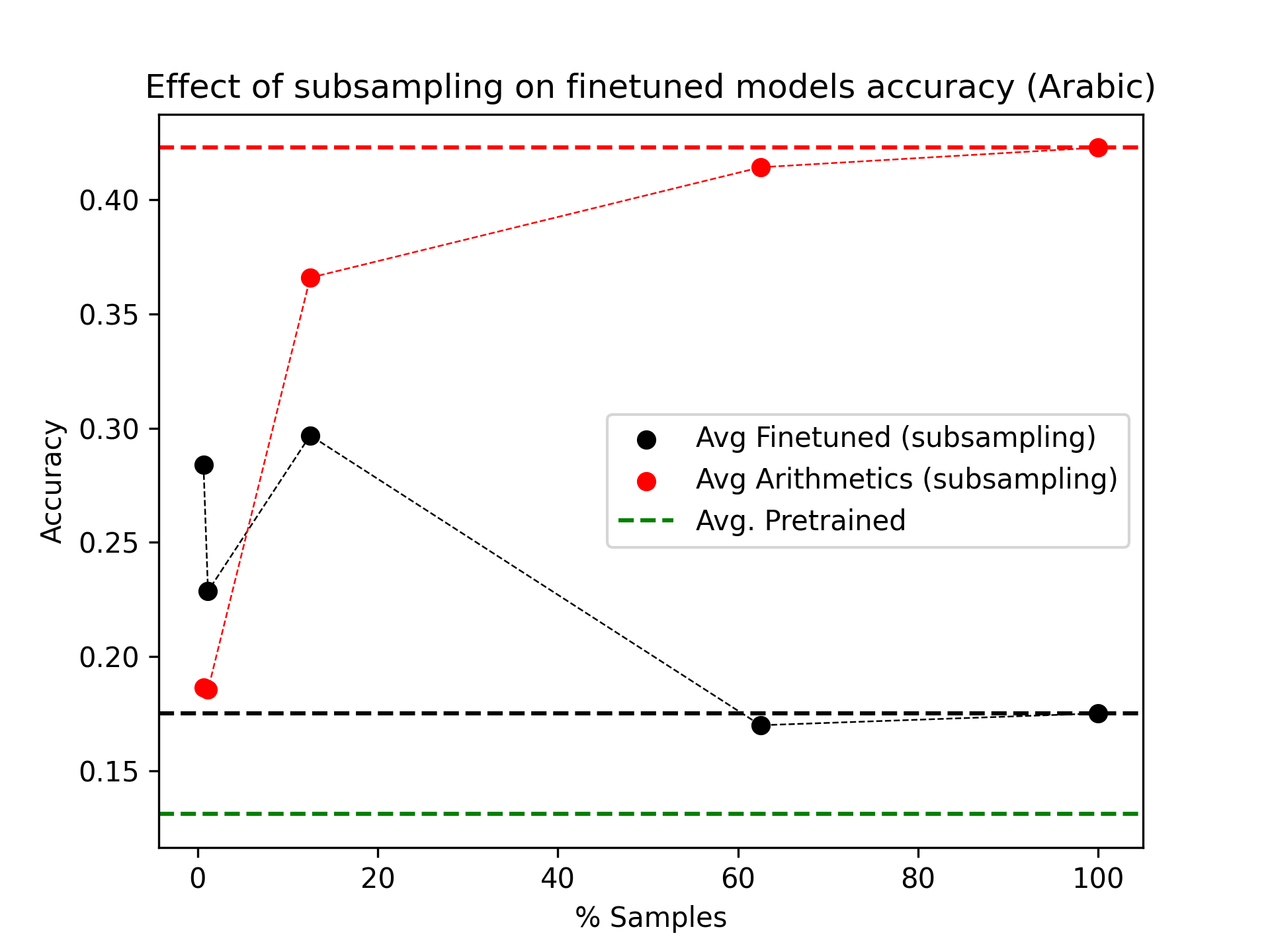}
        \caption{Arabic}
        \label{fig:individual_subsampling_arabic}
    \end{subfigure}
        \begin{subfigure}{0.32\textwidth}
        \centering
        \includegraphics[width=\textwidth]{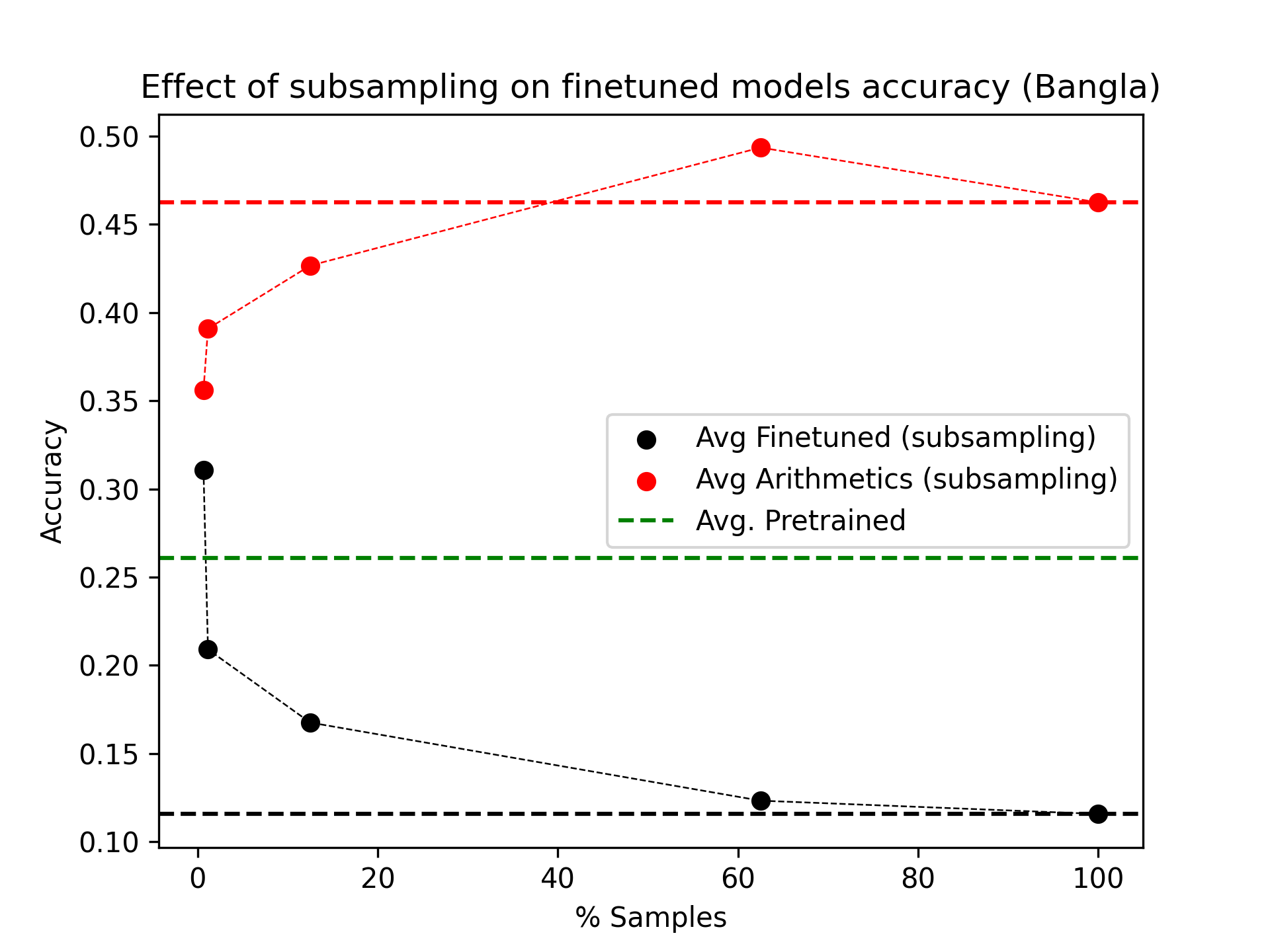}
                \caption{Bangla}

        \label{fig:individual_subsampling_bangla}
    \end{subfigure}
    \begin{subfigure}{0.32\textwidth}
        \centering
        \includegraphics[width=\textwidth]{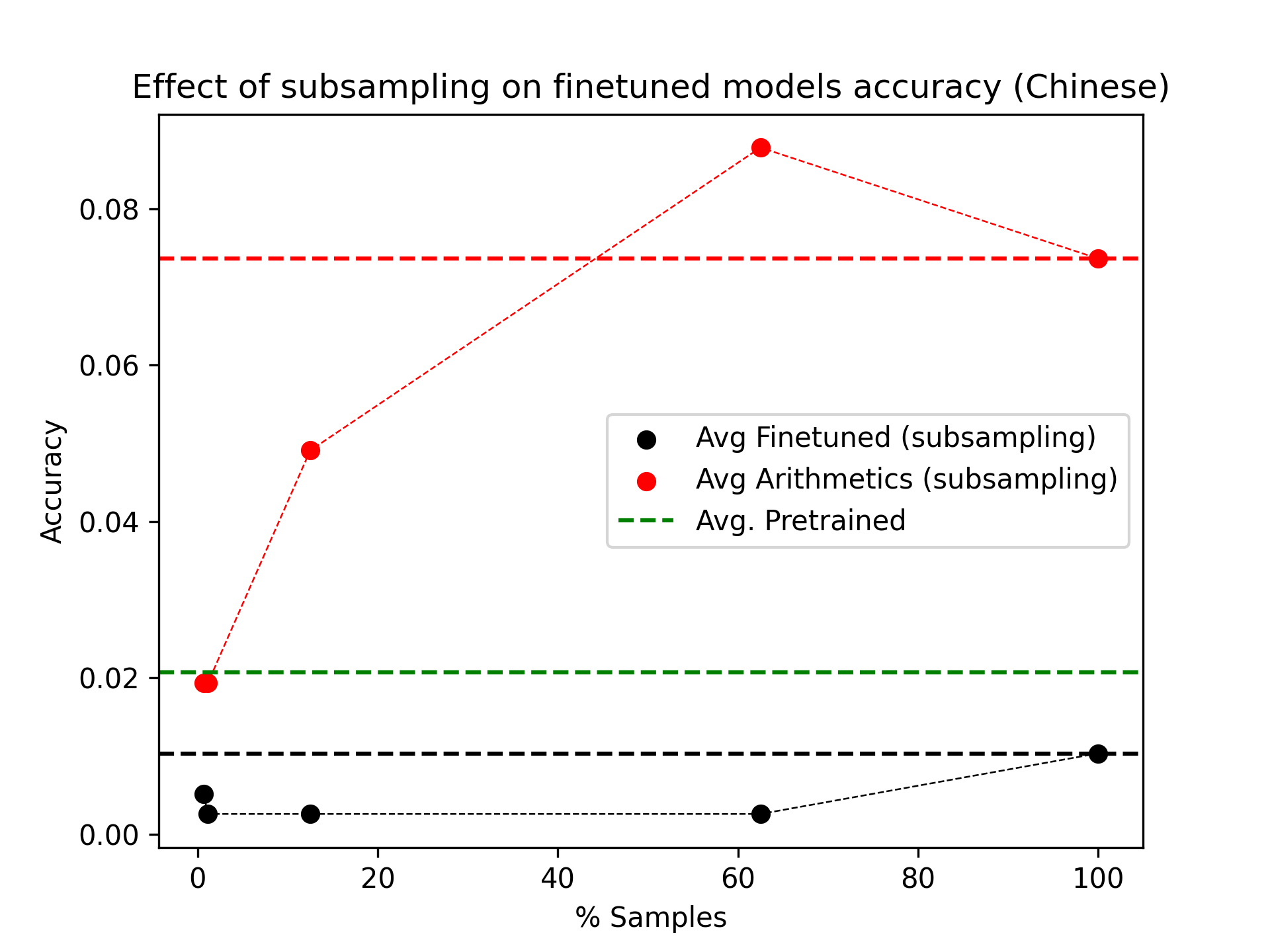}
                \caption{Chinese}

        \label{fig:individual_subsampling_chinese}
    \end{subfigure} \\
    \begin{subfigure}{0.32\textwidth}
        \centering
        \includegraphics[width=\textwidth]{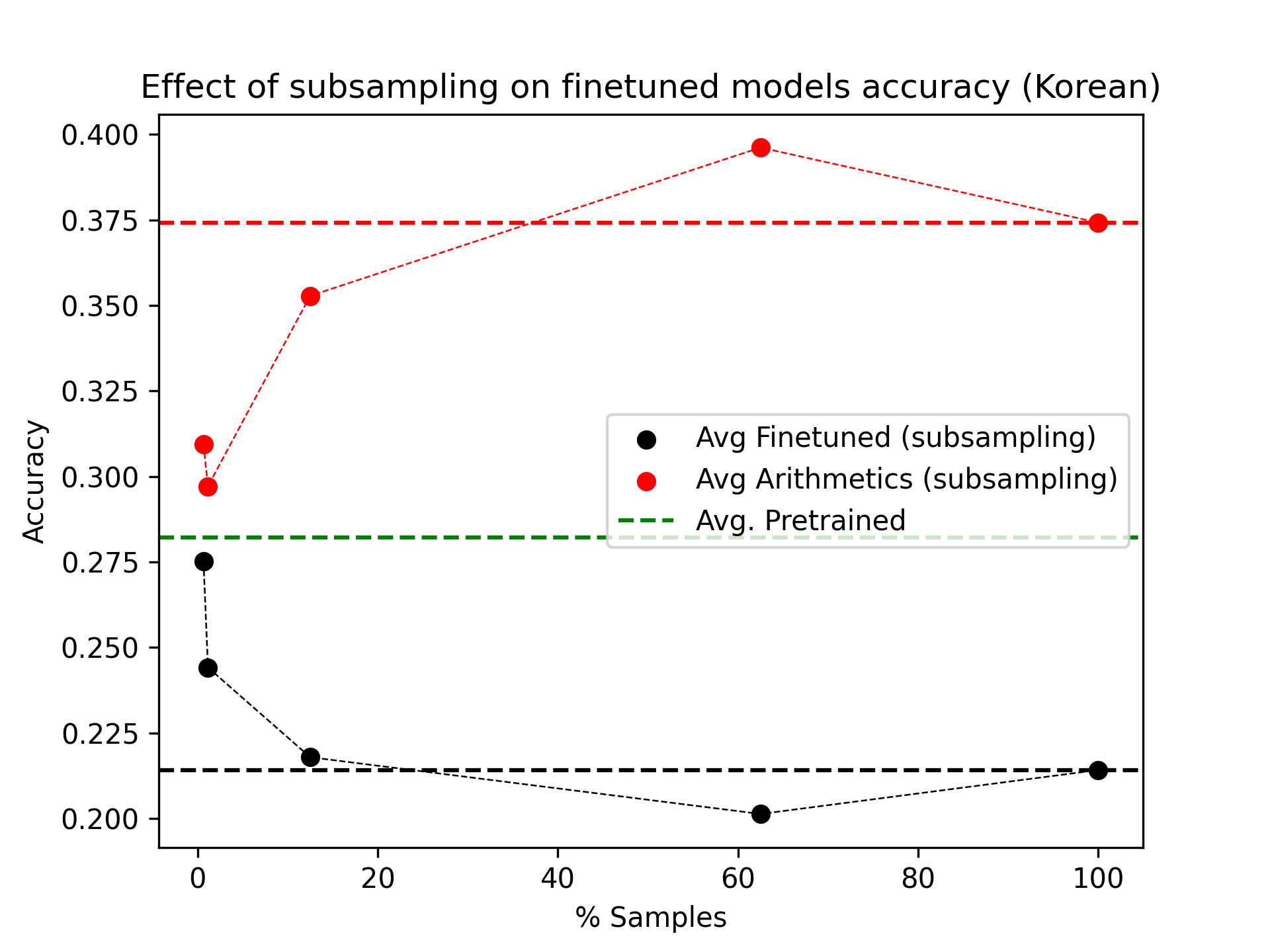}
        \caption{Korean}
        \label{fig:individual_subsampling_korean}
    \end{subfigure}
        \begin{subfigure}{0.32\textwidth}
        \centering
        \includegraphics[width=\textwidth]{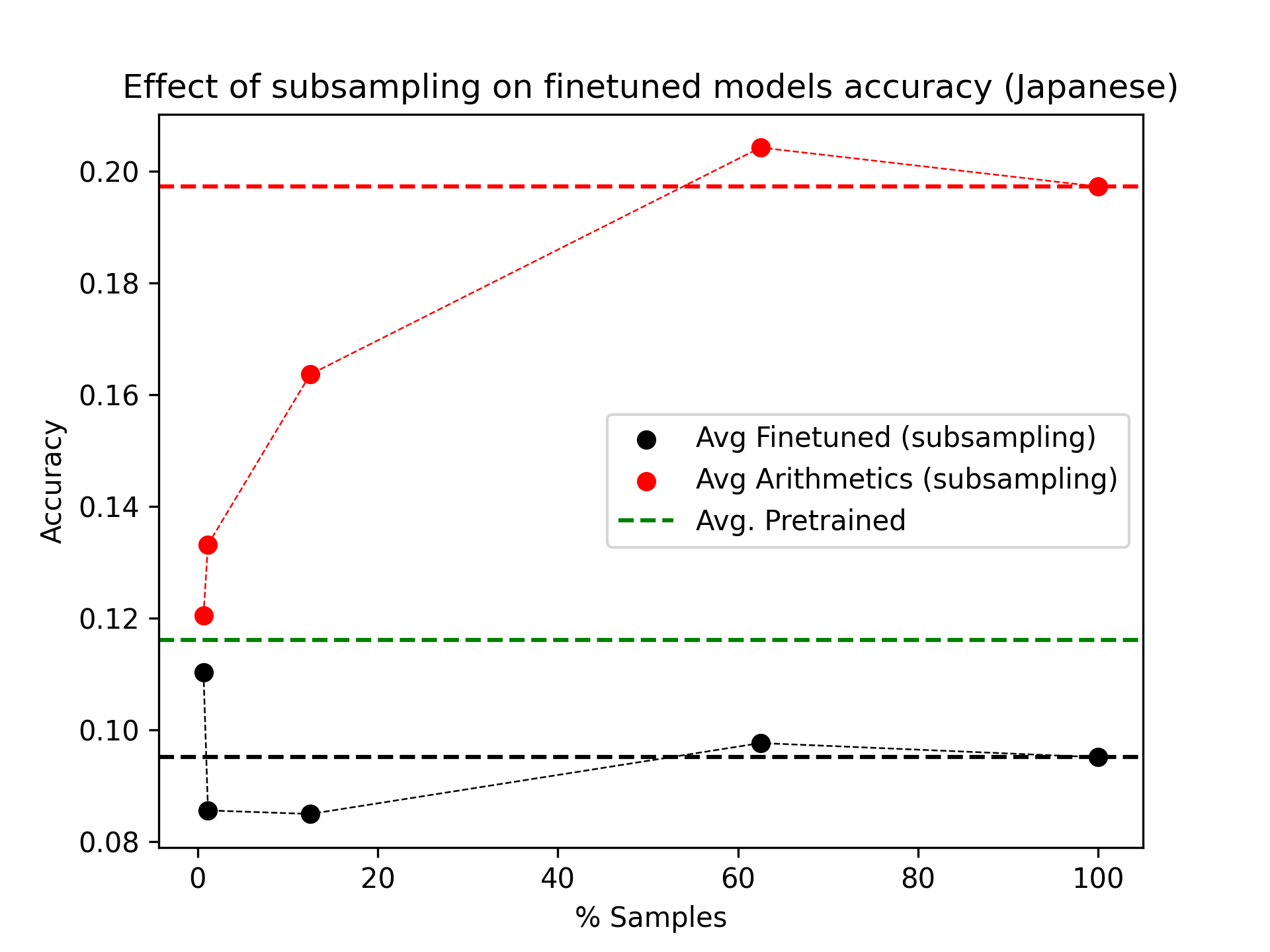}
                \caption{Japanese}

        \label{fig:individual_subsampling_japan}
    \end{subfigure}
    \begin{subfigure}{0.32\textwidth}
        \centering
        \includegraphics[width=\textwidth]{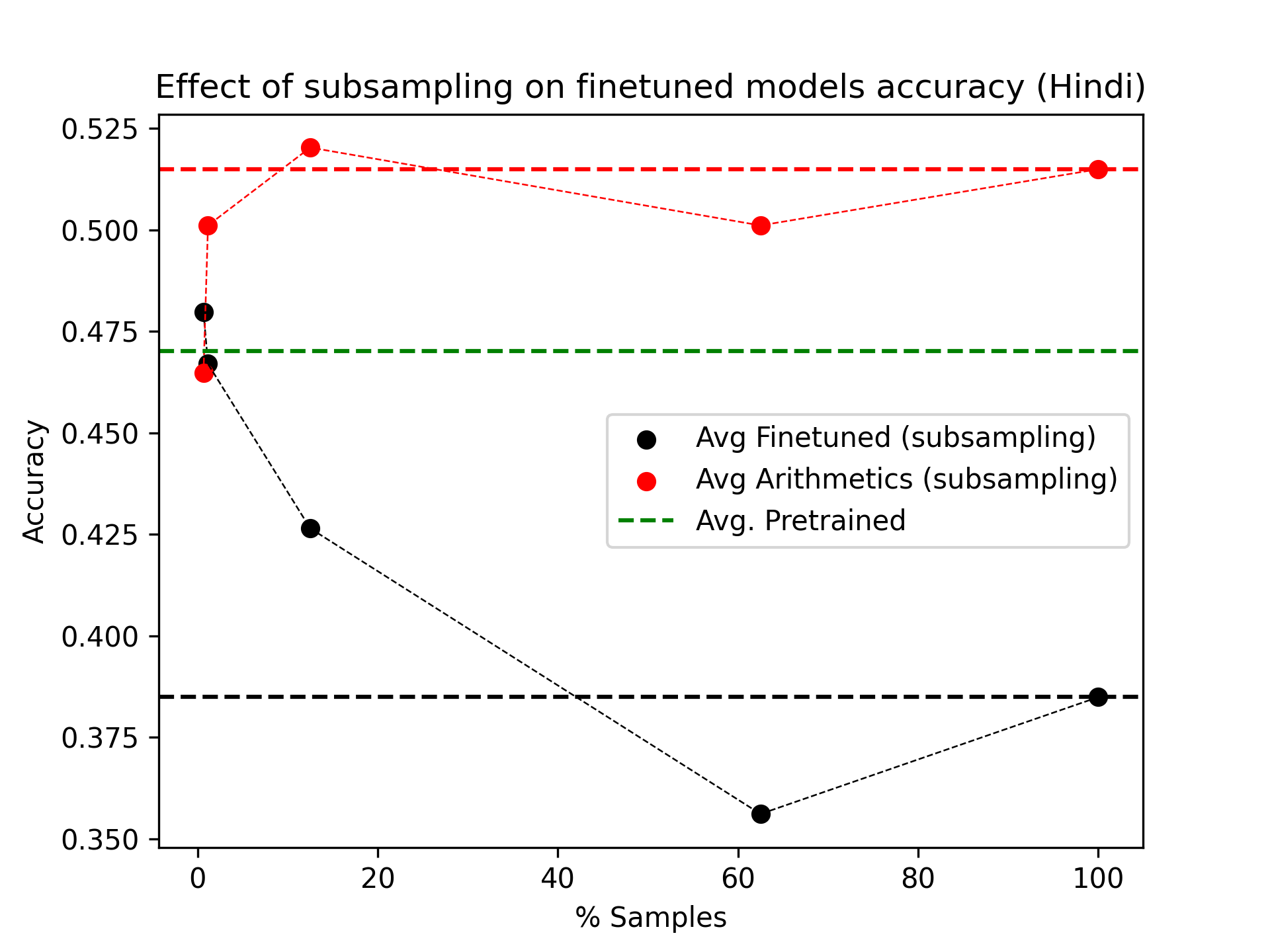}
                \caption{Hindi}

        \label{fig:individual_subsampling_hindi}
    \end{subfigure}
        \begin{subfigure}{0.32\textwidth}
        \centering
        \includegraphics[width=\textwidth]{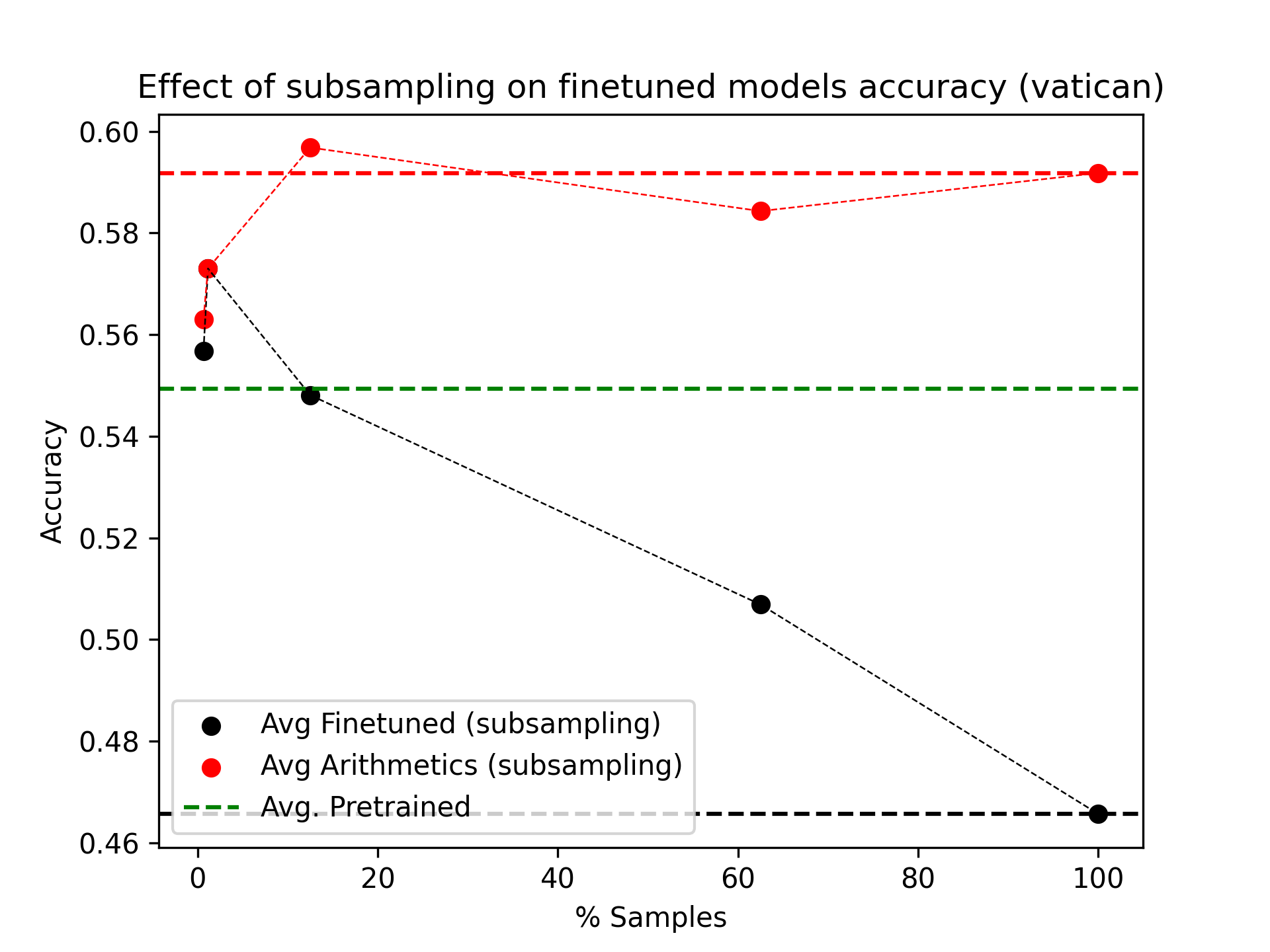}
        \caption{Vatican}
        \label{fig:individual_subsampling_vatican}
    \end{subfigure}
        \begin{subfigure}{0.32\textwidth}
        \centering
        \includegraphics[width=\textwidth]{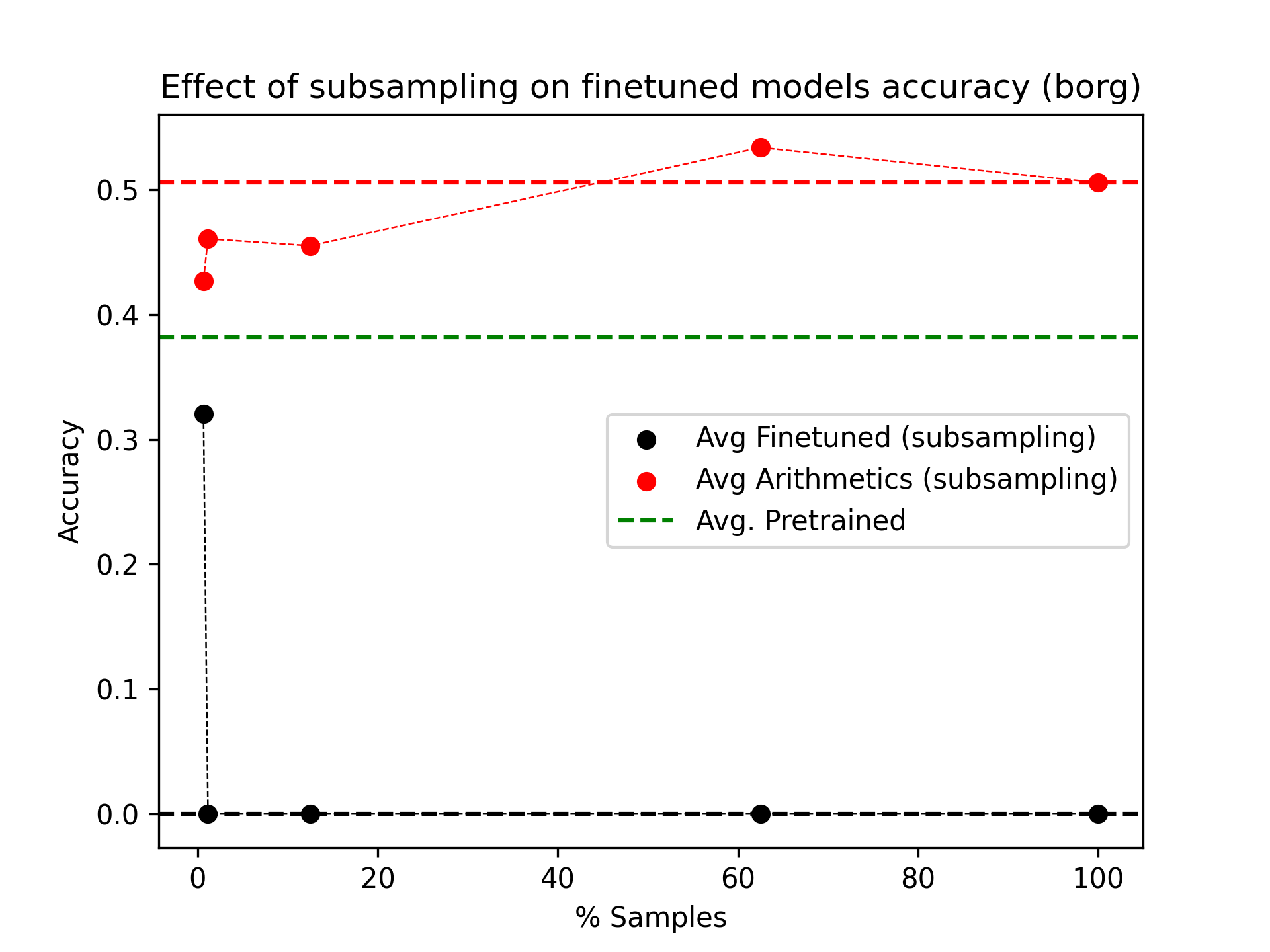}
                \caption{Borg}

        \label{fig:individual_subsampling_borg}
    \end{subfigure}
    \begin{subfigure}{0.32\textwidth}
        \centering
        \includegraphics[width=\textwidth]{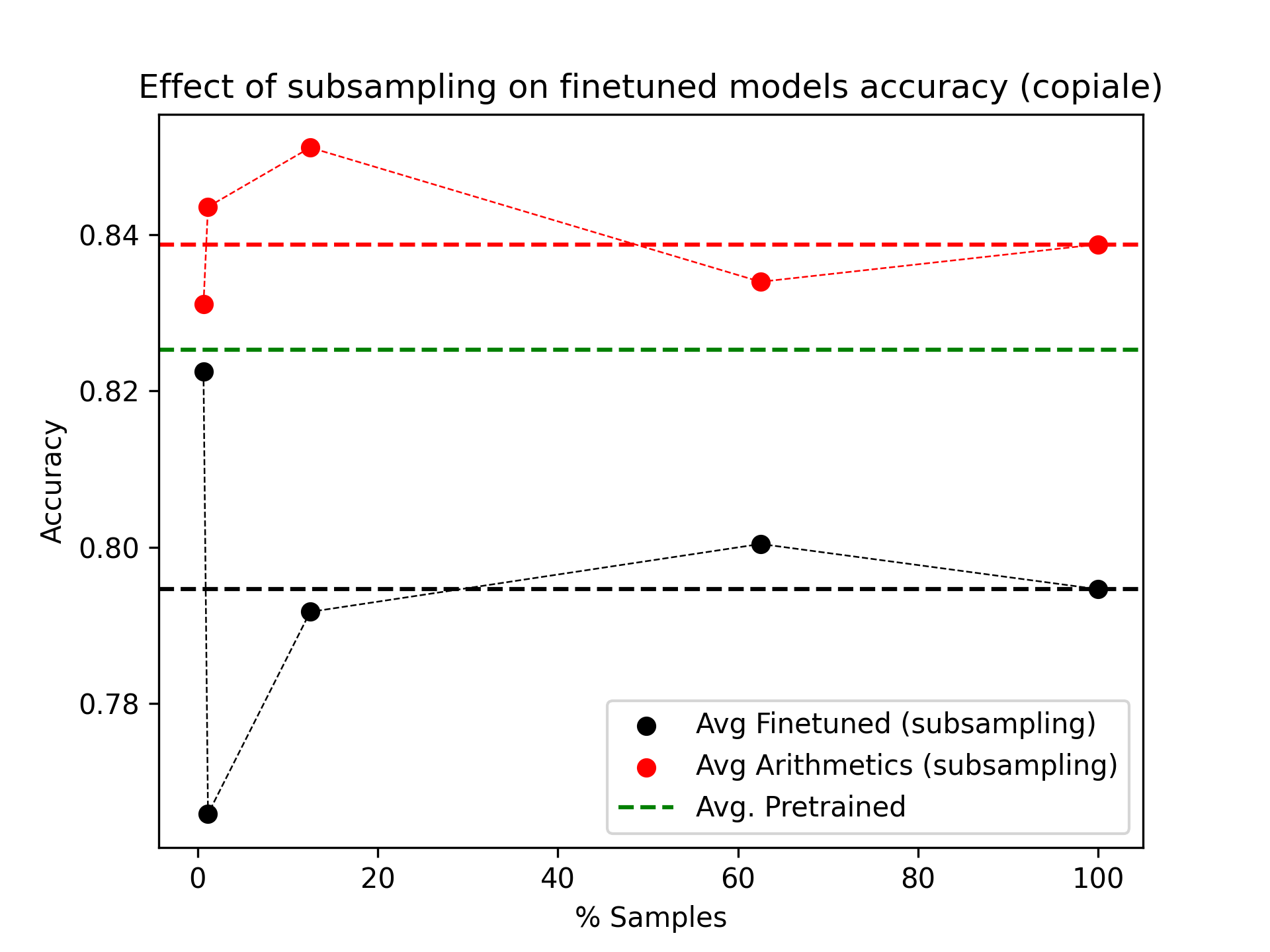}
                \caption{Copiale}

        \label{fig:individual_subsampling_copiale}
    \end{subfigure} 
    \caption{Performance of the different models trained on multilingual and ciphered datasets under different sub-sampling proportions of the original data.}
    \label{fig:languages_individual_detailed}
\end{figure}
\begin{figure*}[t]
    \centering
    \includegraphics[width=0.49\linewidth]{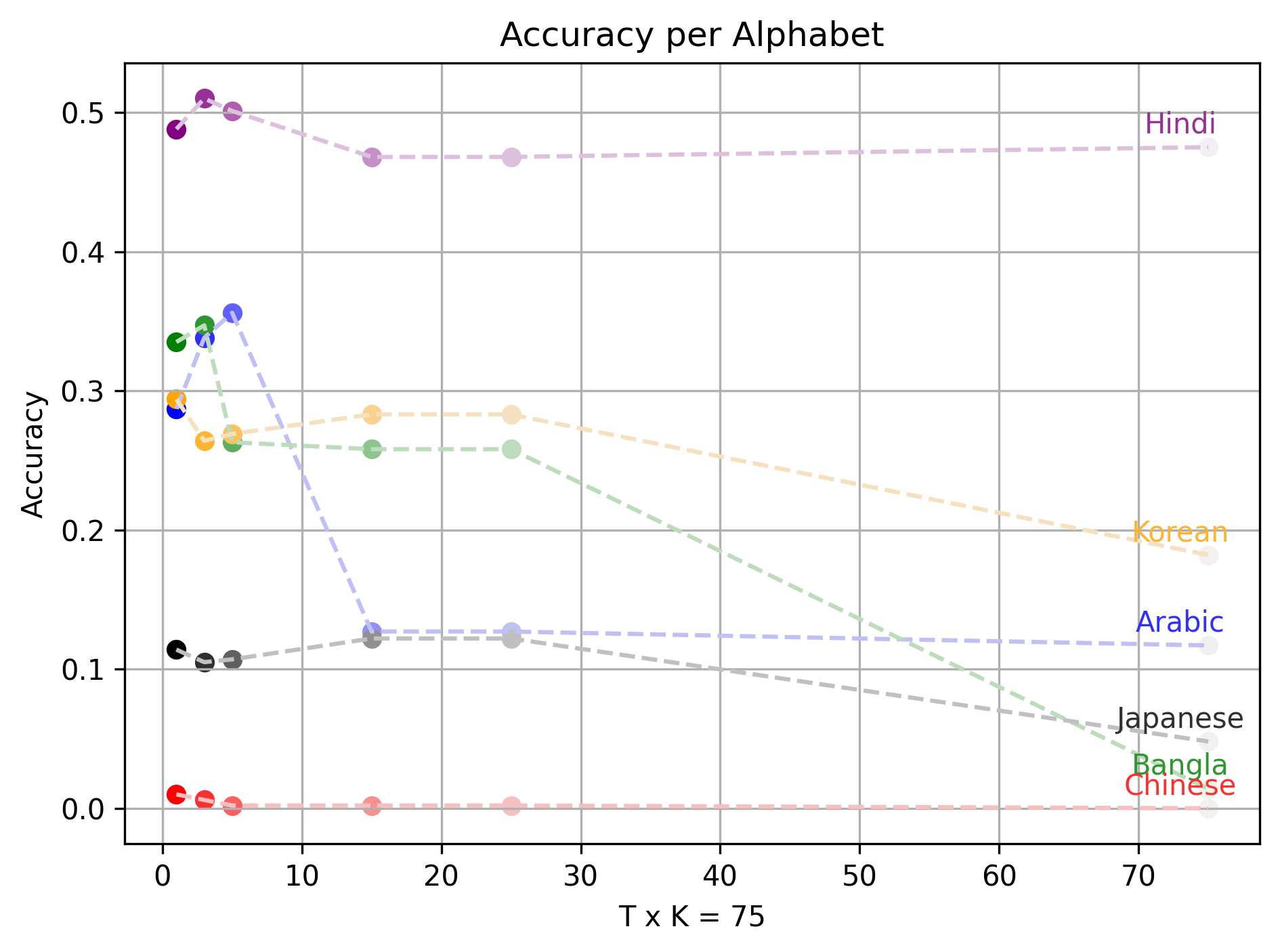}
        \includegraphics[width=0.49\linewidth]{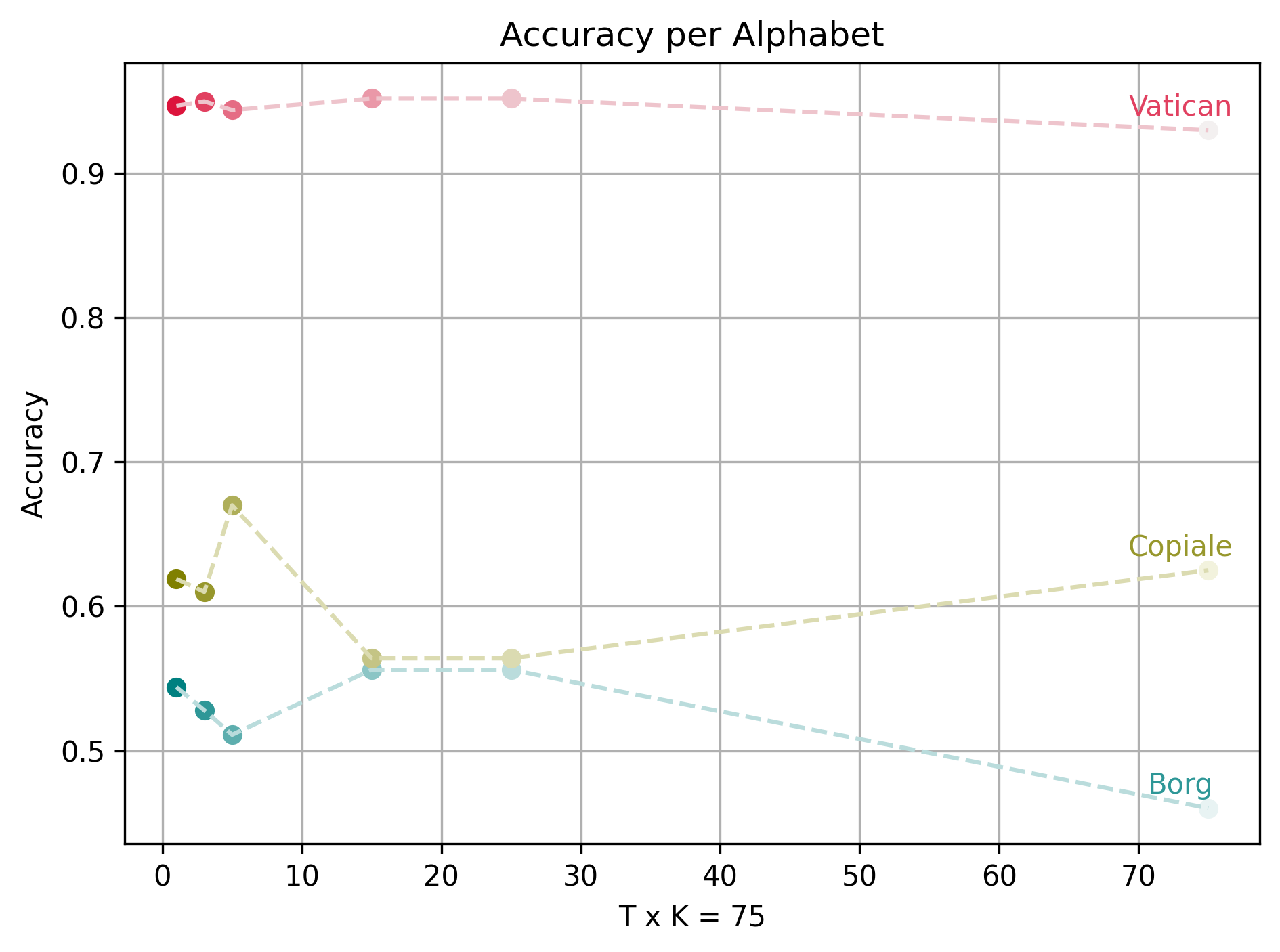}

    \caption{Accuracy of models pre-trained under a different number of epochs ($k$) and distributed rounds ($T$). All models are pre-trained under the same number of forward passes and trained under the same regime and dataset for each new alphabet.}
    \label{fig:performace_per_reptile}
\end{figure*}
As it was previously stated, in this section we aim to address two questions regarding the effectiveness of this method for incorporating new scripts in character recognition systems. First, whether the learnt features of generalist models is independent from the extensiveness of the meta-learnt process or rather the models perform independently of its computing expense. As shown in Figure~\ref{fig:scale_actor}, we examine the average accuracy and standard deviation of fine-tuned models for both non-western scripts and historical ciphers across different sub-sampling percentages. We observe that performance converges upon reaching a critical data threshold (approximately 60\% of $d_n$ for each sub-domain). However, at any sub-sampling percentage, our method consistently outperforms centralized pre-training. In Figure~\ref{fig:languages_individual_detailed} we observe how some alphabets (such as Chinese or Borg) are on a need for task arithmetic in order to converge fast enough to yield any result.

From this analysis, we observe that the experiments in this section provide evidence that task arithmetic becomes more prominent at large pre-training scales. However, it is important to note that even in smaller sub-sampling regimes, the method consistently outperforms a centralized pre-training approach. This suggests that the benefits of task arithmetic extend beyond just large-scale scenarios, offering advantages even when data availability is limited.

Secondly, we inspect up to which degree the number of meta-learning steps $T$ can fatherly improve the results with respect using task arithmetic ($T=1$). In other words, it is explored if there is a limit for generalization on meta-learning regimes. Figure~\ref{fig:performace_per_reptile} shows the achieved accuracy after pre-training on $T = \{1, 3 \dots 75\}$ and $k = \{75, 25 \dots 1\}$. It is ensured to maintain a ratio of $k \times T = 75$ to guarantee fair comparisons. From the computational point of view, every trained model performed the same amount of backward / forward passes on $D$, but distributed under a different amount of meta-learning steps ($T$). From the analysis conducted in Figure~\ref{fig:performace_per_reptile} we conclude that task arithmetic is not all you need, but rather a close version of the ideal case of meta-learning during a few distributed rounds. After performing many meta-learning steps (i.e. few steps on $U^k$ since $T=75$ and $k=1$) the performance decays due to an over-specialization of the model on the original source domains.

In Table~\ref{tab:ablation}, we present an ablation study of different model combinations to inspect its effect upon learning new language. This is, leaving-one-out to study its \textit{individual contribution}, using task vectors associated with \textit{handwritten recognition} tasks, testing the contribution of language modeling through ignoring \textit{non-English} recognition tasks and whether the historical nature of some documents may positively impact the recognition of ciphers. Final accuracy is presented for every new script learnt during 30 stochastic optimization epochs. Some of the results noted in this ablation are the importance of ignoring handwritten tasks for learning MLT sub-sets, which makes sense as non-English datasets in this study contain solely scene images. However, in such cases, handwritten does not consistently mislead the final model. This discrepancy is not found on learning ciphers, where in some cases such as in the Copiale dataset, using scene images from MLT leads to slightly worse performance.

\paragraph{\textbf{Computational Resources}}  
The computational resources used for this project are relatively modest, as we advocate for smaller and simpler architectures that facilitate exploration. Each individual training run is conducted on a single GeForce RTX 3090 GPU (23.70 GiB of vRAM). Training on small datasets (1K to 10K samples) takes approximately 5 hours for 30 epochs and over 10 hours for 75 epochs. For larger datasets (100K to 1M samples), optimization can range from 3 to 5 days for 75 epochs, which applies specifically to HierText and TextOCR training (see Section~\ref{sec:data}).  

To parallelize essential experiments, we utilized a maximum of 8 GPUs (4 RTX 3090 and 4 A40). On A40 GPUs, we maintained the same batch size to ensure reproducibility and fair comparisons, disregarding the additional 20 GiB of unused vRAM. This setup enables reproducible experiments that are accessible to many medium-sized laboratories. Considering failure cases, hyperparameter tuning, and discarded experiments, our logs report a total of approximately 3,000 GPU hours.  
\begin{table*}[t]
    \centering
     \resizebox{\textwidth}{!}{\begin{tabular}{llllllll|lllllllll|}
\cline{9-17}
                                                                                                          &            &     &     &          &      &     &      & \multicolumn{9}{l|}{Target Alphabet}                                                  \\ \cline{2-17} 
\multicolumn{1}{l|}{}                                                                                     & \multicolumn{7}{l|}{Task Vector}                      & \multicolumn{6}{l|}{Scene}                              & \multicolumn{3}{l|}{Cipher} \\ \cline{2-17} 
\multicolumn{1}{l|}{}                                                                                     & Esposalles & IAM & GW  & Parzival & COCO & MLT & TOCR & Arab & Chi & Ban & Hin & Kor & \multicolumn{1}{l|}{Jap} & Borg  & Copiale  & Vatican  \\ \hline
\multicolumn{1}{|l|}{\multirow{7}{*}{\begin{tabular}[c]{@{}l@{}}Leave-One-Out\end{tabular}}}  & \xmark        & \cmark& \cmark& \cmark     & \cmark & \cmark& \cmark & .386  & .090 & .477 & .530 & .361 & \multicolumn{1}{l|}{.204} & .544   & .944      & .616      \\
\multicolumn{1}{|l|}{}                                                                                    & \cmark       & \xmark & \cmark& \cmark     & \cmark & \cmark& \cmark & .385  & .081 & .472 & .485 & .403 & \multicolumn{1}{l|}{.201} & .550   & .948      & \color{green}\textbf{.629}\color{black}      \\
\multicolumn{1}{|l|}{}                                                                                    & \cmark       & \cmark& \xmark & \cmark     & \cmark & \cmark& \cmark & .382  & .077 & .471 & .508 & .395 & \multicolumn{1}{l|}{.200} & .550   & .950      & .612      \\
\multicolumn{1}{|l|}{}                                                                                    & \cmark       & \cmark& \cmark& \xmark      & \cmark & \cmark& \cmark & .384  & .082 & .471 & .502 & .379 & \multicolumn{1}{l|}{.208} & .573   & .937      & .616      \\
\multicolumn{1}{|l|}{}                                                                                    & \cmark       & \cmark& \cmark& \cmark     & \xmark  & \cmark& \cmark & .384  & .089 & .469 & .510 & .397 & \multicolumn{1}{l|}{.206} & .556   & .943      & .624      \\
\multicolumn{1}{|l|}{}                                                                                    & \cmark       & \cmark& \cmark& \cmark     & \cmark & \xmark & \cmark & .292  & .023 & .378 & .529 & .297 & \multicolumn{1}{l|}{\color{red}\textbf{.128}\color{black}} & .533   & \color{green}\textbf{.951}\color{black}      & .612      \\
\multicolumn{1}{|l|}{}                                                                                    & \cmark       & \cmark& \cmark& \cmark     & \cmark & \cmark& \xmark  & .360  & .080 & .452 & .530 & .397 & \multicolumn{1}{l|}{.204} & \color{red}\textbf{.500}\color{black}   & \color{red}\textbf{.934}\color{black}      & .614      \\ \hline
\multicolumn{1}{|l|}{\multirow{2}{*}{\begin{tabular}[c]{@{}l@{}}Handwritten \\ Ablation\end{tabular}}} & \cmark       & \cmark& \cmark& \cmark     & \xmark  & \xmark & \xmark  & .251  & .020 & .369 & .468 & \color{red}\textbf{.275}\color{black} & \multicolumn{1}{l|}{\color{red}\textbf{.128}\color{black}} & .522   & .940      & .596      \\
\multicolumn{1}{|l|}{}                                                                                    & \xmark        & \xmark & \xmark & \xmark      & \cmark & \cmark& \cmark & \color{green}\textbf{.462}\color{black}  & \color{green}\textbf{.117}\color{black} & \color{green}\textbf{.486}\color{black} & .514 & \color{green}\textbf{.430}\color{black} & \multicolumn{1}{l|}{\color{green}\textbf{.240}\color{black}} & \color{green}\textbf{.578}\color{black}   & .939      & .617      \\ \hline
\multicolumn{1}{|l|}{\multirow{2}{*}{\begin{tabular}[c]{@{}l@{}}Non-Eng\\ Ablation\end{tabular}}}     & \cmark       & \xmark & \xmark & \cmark     & \xmark  & \xmark & \xmark  & \color{red}\textbf{.236}\color{black}  & .027 & \color{red}\textbf{.332}\color{black} & \color{red}\textbf{.461}\color{black} & .285 & \multicolumn{1}{l|}{\color{red}\textbf{.128}\color{black}} & .516   & .941      & .601      \\
\multicolumn{1}{|l|}{}                                                                                    & \xmark        & \cmark& \cmark& \xmark      & \cmark & \cmark& \cmark & .409  & .094 & .469 & .514 & .411 & \multicolumn{1}{l|}{.215} & .556   & .945      & .604      \\ \hline
\multicolumn{1}{|l|}{\multirow{2}{*}{Historical}}                                                         & \cmark       & \xmark & \cmark& \cmark     & \xmark  & \xmark & \xmark  & .278  & \color{red}\textbf{.016}\color{black} & .338 & .463 & .288 & \multicolumn{1}{l|}{.135} & .522   & .943      & \color{red}\textbf{.586}\color{black}      \\
\multicolumn{1}{|l|}{}                                                                                    & \xmark        & \cmark& \xmark & \xmark      & \cmark & \cmark& \cmark & .401  & .098 & .478 & \color{green}\textbf{.549}\color{black} & .411 & \multicolumn{1}{l|}{.220} & .556   & .946      & .622      \\ \hline
\end{tabular}
}   
    \caption{Ablation study containing the accuracy of models fine-tuned after a task arithmetic procedure. \color{green}Best \color{black}  and \color{red}worst \color{black}  accuracy is reported for every script.}
    \label{tab:ablation}
\end{table*}

For practitioners aiming to fine-tune a pre-trained model, the estimated computational requirements are approximately 1 week of pre-training on a single RTX 3090 GPU, followed by around 5 days for training on sub-domains in a single-threaded, sequential manner. Model averaging takes roughly 1 minute on a CPU, while transfer learning requires about 5 hours. This total time can be significantly reduced by leveraging the distributed nature of our method, particularly by utilizing multiple GPUs in parallel during the sub-domain specialization stage.

\section{Conclusions}
\label{sec:conc}

In this paper, we have empirically demonstrated the significant potential of task arithmetic and meta-learning in the context of adapting to low-resource unseen scripts in intelligent reading systems. Our findings provide evidence that training models using task arithmetic improves out-of-domain generalization. Moreover, we present strong evidence showing that this approach can achieve up to three times the performance of our baseline and classical fine-tuning methods, which is highly promising for tasks and applications where an OCR system plays a central role.

In summary, as a take-home message for OCR generalization, we propose that practitioners begin by creating a subsample of the pre-training dataset, which is agnostic of specific modalities or organizational boundaries, but shares a common high-resource alphabet. This approach enables the training of multiple expert models on distributed subsets, which collectively encompass the same diversity as the original dataset $D$. The aggregated expertise of these models is particularly effective in out-of-domain scenarios, allowing the averaged model to better handle unannotated documents with significant distribution shifts. Furthermore, if these documents contain alphabets that differ from those in the training data, a brief period of fine-tuning with minimal annotations and a few epochs can rapidly adapt the model to the new conditions.

The paper shows that task arithmetic and distributed training enable robust OCR across low-resource alphabets and historical ciphered texts, outperforming classical fine-tuning while retaining high in-domain accuracy.

\paragraph{\textbf{Ethical Concerns}}  
In line with our conclusions, we also wish to address the ethical concerns that may arise from using our method. Since the performance of our approach is inherently constrained by the capabilities of the distributedly trained models, practitioners might be tempted to scale each training node indefinitely, aiming to enhance overall scalability in proportion to the number of nodes. However, our findings (see Figure~\ref{fig:scale_actor}) indicate that global performance tends to plateau beyond a critical data threshold. This suggests that excessive scaling may lead to diminishing returns rather than sustained improvements. Therefore, we encourage future research to carefully assess the necessary scaling factor to prevent unnecessary computational expansion and the associated energy costs.

\section*{Acknowledgements}

This work has been partially supported by the Spanish project PID2021-126808OB-I00, Ministerio de Ciencia e Innovación, the Departament de Cultura of the Generalitat de Catalunya, and the CERCA Program / Generalitat de Catalunya. Adrià Molina is funded with the PRE2022-101575 grant provided by MCIN / AEI / 10.13039 / 501100011033 and by the European Social Fund (FSE+).

\clearpage
\newpage
\bibliographystyle{splncs04}
\bibliography{001_main}

\end{document}